\documentclass[acmtog,nonacm,authorversion]{acmart}

\AtBeginDocument{%
  \providecommand\BibTeX{{%
    \normalfont B\kern-0.5em{\scshape i\kern-0.25em b}\kern-0.8em\TeX}}}

\setcopyright{acmcopyright}
\acmJournal{TOG}
\acmYear{2023}
\acmVolume{00}
\acmNumber{0}
\acmArticle{000}
\acmMonth{12}
\acmDOI{10.1145/xxxxxx.xxxxxx}

\usepackage{bm}

\usepackage{cancel}

\usepackage[textsize=tiny]{todonotes}

\usepackage{graphicx}
\usepackage[labelformat=simple]{subcaption}
\usepackage{amsmath}
\usepackage{float}
\usepackage{soul}
\usepackage{mathrsfs}

\newcommand{\vect}[1]{\bm{#1}}

\usepackage{multirow} 
\usepackage{makecell} 
\usepackage{utfsym} 
\usepackage{hyphenat} 

\usepackage{booktabs} 
\usepackage[normalem]{ulem}
\usepackage{enumitem}

\usepackage[ruled]{algorithm2e} 

\SetAlFnt{\small}
\SetAlCapFnt{\small}
\SetAlCapNameFnt{\small}
\SetAlCapHSkip{0pt}

\usepackage{cases}
\usepackage{subcaption}
\usepackage{xspace}
\newcommand{\ie}{\textsl{i.e.}~}

\newcommand{\keep}[1]{}
\newcommand{\old}[1]{}

\DeclareMathOperator*{\argmin}{ {arg\,min} }

\newcommand{\loss}{\ensuremath{\mathcal{L}}}

\newcommand{\worldmodel}{\ensuremath{\mathcal{W}}}

\newcommand{\encoder}{\ensuremath{\mathcal{E}}}
\newcommand{\decoder}{\ensuremath{\mathcal{D}}}
\newcommand{\codebook}{\ensuremath{\mathcal{B}}}

\newcommand{\fig}{Figure{}~}
\newcommand{\eqn}{Equation{}~}

\newcommand{\stt}{\ensuremath{\vect{s}}}
\newcommand{\act}{\ensuremath{\vect{a}}}
\newcommand{\policy}{\ensuremath{{\vect{\pi}}}}

\newcommand{\latent}{\ensuremath{\vect{v}}}
\newcommand{\latentseq}{\ensuremath{\vect{V}}}
\newcommand{\latentcode}{\ensuremath{\vect{z}}}
\newcommand{\latentvq}{\ensuremath{\vect{z}}}
\newcommand{\latentvqseq}{\ensuremath{\vect{Z}}}
\newcommand{\LatentSpace}{\ensuremath{\mathcal{Z}}}

\newcommand{\vqidx}{\ensuremath{{I}}}
\newcommand{\vqidxseq}{\ensuremath{{S}}}
\newcommand{\midlatent}{\ensuremath{\vect{u}}}

\newcommand{\task}{\ensuremath{\vect{g}}}
\newcommand{\tarmo}{\ensuremath{\vect{M}}}

\newcommand{\vqvae}{VQ\hyp{}VAE}
\newcommand{\frameworkname}{MoConVQ}
\newcommand{\controlGPT}{MoConGPT}

\newcommand{\numcode}{\ensuremath{{N}_{\codebook}}}

\newcommand{\textPrompt}[1]{{\textit{\dashuline{#1}}}}

\begin{document}

\title{\frameworkname{}: Unified Physics-Based Motion Control via Scalable Discrete Representations}

\author{Heyuan Yao}
\email{heyuanyao@pku.edu.cn}
\orcid{0000-0002-6168-6777}

\author{Zhenhua Song}
\email{songzhenhua@stu.pku.edu.cn}
\orcid{0000-0001-5139-209X}

\author{Yuyang Zhou}
\email{yuyangzhou2002@gmail.com}
\orcid{0009-0000-3355-0306}

\author{Tenglong Ao}
\email{aubrey.tenglong.ao@gmail.com}
\orcid{0000-0002-7418-1014}
\affiliation{%
  \institution{Peking University}
  \city{Beijing}
  \country{China}
}

\author{Baoquan Chen}
\email{baoquan@pku.edu.cn}
\orcid{0000-0003-4702-036X}  

\author{Libin Liu}
\authornote{corresponding author}
\email{libin.liu@pku.edu.cn}
\orcid{0000-0003-2280-6817}
\affiliation{%
  \institution{Peking University}
  \city{Beijing}
  \country{China}
}
\affiliation{%
  \institution{National Key Lab of General AI}
  \city{Beijing}
  \country{China}
}

\renewcommand{\shortauthors}{Yao, Song, Zhou, Ao, Chen, and Liu}

\begin{abstract}

    In this work, we present \frameworkname{}, a novel unified framework for physics-based motion control leveraging scalable discrete representations. Building upon vector quantized variational autoencoders (VQ-VAE) and model-based reinforcement learning, our approach effectively learns motion embeddings from a large, unstructured dataset spanning tens of hours of motion examples. The resultant motion representation not only captures diverse motion skills but also offers a robust and intuitive interface for various applications. We demonstrate the versatility of \frameworkname{} through several applications: universal tracking control from various motion sources, interactive character control with latent motion representations using supervised learning, physics-based motion generation from natural language descriptions using the GPT framework, and, most interestingly, seamless integration with large language models (LLMs) with in-context learning to tackle complex and abstract tasks.

\end{abstract}
    


\newlength{\twosubht}
\setlength{\twosubht}{2.85cm}
\begin{teaserfigure}
\subcaptionbox{Motion Tracking}{%
  \includegraphics[height=\twosubht]{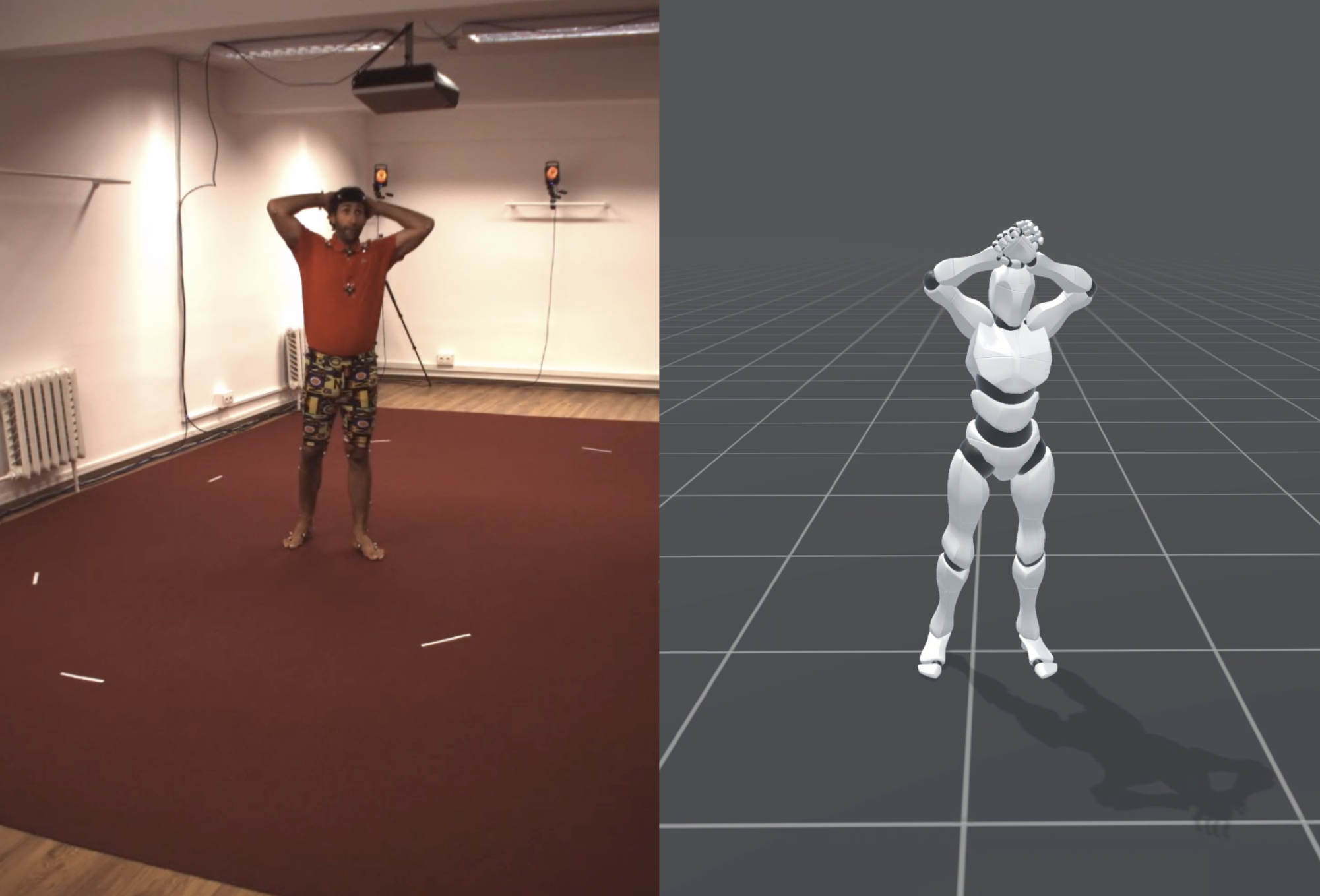}%
}\hfill
\subcaptionbox{Interactive Control}{%
  \includegraphics[height=\twosubht]{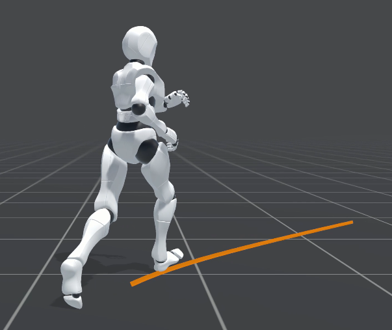}%
}\hfill
\subcaptionbox{Text2Motion}{%
  \includegraphics[height=\twosubht]{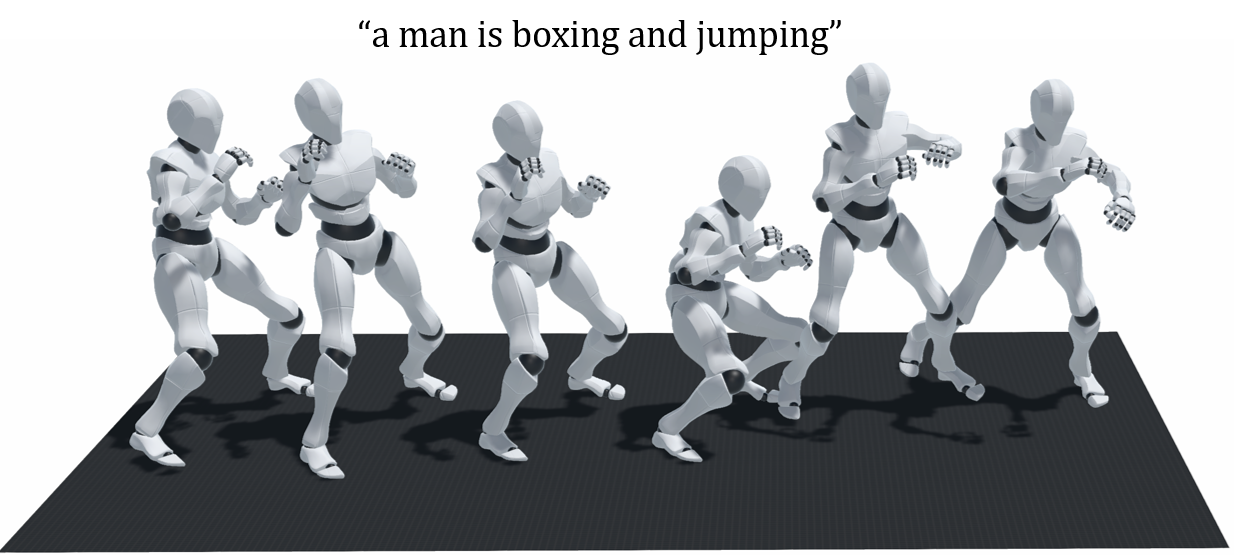}%
}\hfill
\subcaptionbox{LLM Integration}{%
  \includegraphics[height=\twosubht]{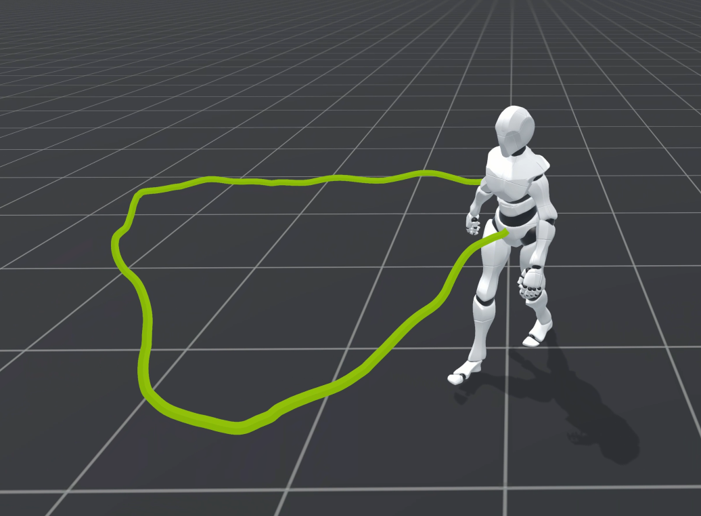}%
}
\caption{We present a method for learning discrete motion representation from a large-scale unstructured motion dataset for physics-based characters. The framework allows various applications, including those shown in this figure, to be accomplished in a unified fashion.}
\end{teaserfigure}

\maketitle

\section{Introduction}

Physics-based motion control allows a character to interact with a physically simulated environment. It has the potential to generate realistic character animation and provide natural responses to environmental changes and perturbations. Such abilities are crucial for digital humans and embodied intelligent agents. However, due to the complexity of full-body human dynamics, realizing control over diverse and agile human motion has been a longstanding challenge. 

Many recent studies in this field draw inspiration from deep learning and generative models. They combine various generative models with deep reinforcement learning (DRL), resulting in generative, physics-based motion representations \cite{PengASELargescaleReusableAdversarial2022,Won2022ConditionalVAE,ControlVAE}. When trained on a set of example motions, these generative control strategies effectively capture the motion skills within that dataset, and allow downstream tasks to be effectively completed using the latent representations. Despite these successes, there are still notable limitations in these approaches. For instance, the datasets used in these studies typically comprise only several minutes or tens of minutes of data with limited motion diversity. It is nontrivial to scale up a DRL framework to accommodate hours or even more diverse motion data. The latent motion representation often lacks explicit meaning, making it challenging to select desired motions from the latent space. To utilize a specific motion, a dedicated reward function is often required to guide a policy toward that particular movement, and extra DRL training is often needed to train the policy.

In this paper, we attempt to address these problems. Particularly, we focus on letting a physics-based character grasp a diverse set of motion skills from a large dataset of tens of hours of example motions, and providing a unified and intuitive interface for various downstream tasks. To this end, we address two  primary questions: (a) what is an effective motion representation for controlling a diverse range of motions? and (b) how to efficiently learn such a representation from a large volume of data?

For the representation of motion, we draw inspiration from recent studies that utilize discrete, or \emph{vector quantized}, latent representations, which have achieved high-quality generation in domains such as images \cite{rombach2022latentdiffusionimage}, music \cite{dhariwal2020jukebox}, and motions \cite{Ao2023GestureDiffuCLIP, jiang2023motiongpt}. We formulate our generative motion control using a vector quantized variational autoencoder (VQ-VAE) framework. We argue that this discrete motion control representation is not only capable of accommodating a diverse range of motions but also serves as a robust and intuitive interface for downstream applications.

For training the framework, we note that several recent works \cite{fussellSuperTrackMotionTracking2021,ControlVAE} have utilized model-based reinforcement learning techniques to achieve efficient learning of complex human actions, particularly generative control policies from a set of motions \cite{ControlVAE}. These methods learn a \emph{world model} to approximate the black-box rigid body dynamics, which offers a differentiable path that allows the gradient of the training objective to directly update the control model. We believe that this property of model-based RL makes it suitable for training large, complex generative neural networks. Thus, we adopt a similar model-based RL approach for training our framework, which in practice allows for efficient training on datasets spanning several tens of hours of motion examples

We refer to this scalable framework as \emph{\frameworkname}. It serves as a unified motion control framework, which we demonstrate with several applications including:
\begin{itemize}
    \item Acting as a universal tracking controller to track motions from various sources, such as kinematic motion synthesis and monocular 3D human pose estimation; 
    \item Serving as a versatile latent space within which a high-level policy can be trained using supervised learning for interactive control of the character;
    \item Functioning as a generative motion prior when integrated with the GPT (Generative Pretrained Transformer) framework, supporting high-quality physics-based motion generation from natural language descriptions; and 
    \item Providing an efficient interface for integration with Large Language Models (LLMs), which enables an LLM to learn to use the motions via in-context learning.
\end{itemize}

In summary, our contribution includes:
\begin{itemize}
    \item We propose a unified framework for physics-based motion generation that can scale up to a very large dataset. This framework generates high-quality physics-based motions and can generalize across a broad range of motions. To the best of our knowledge, this is the first physics-based generative motion control model trained on a dataset of such size and diversity.
    
    \item We showcase several essential tasks and applications using our framework. Notably, our text-conditional GPT generator, termed \emph{T2M-\controlGPT{}}, is the first physics-based approach to address such a general text-to-motion synthesis task.
     
    \item We introduce a novel integration with large language models that allows an LLM to understand human motions and use them in a simulated environment. We believe this opens up the possibility of creating intelligent simulated avatars through the aid of LLMs, harnessing their advanced reasoning capabilities to tackle complex and abstract tasks.
\end{itemize}

\section{Related Work}

Our work spans several domains, including skeleton animation, reinforcement learning, and generative models. We review the most relevant works, with a particular focus on physics\hyp{}based motion generation models. For a more in-depth understanding, we direct readers to the surveys presented by \citet{SurveyRL} and \citet{SurveySkeletonHumanAnimation}.

\subsection{Physics-Based Motion Controllers}

Generating realistic and interactive locomotion for simulated characters is a longstanding challenge in computer animation. Early research has been focus on designing motion controllers based on domain knowledge, such as leveraging abstract models to assist the character in maintaining balance, enabling stable walking \cite{yinSIMBICONSimpleBiped2007,constraintawlking2008,Lee2010_Datadriven} and stair-stepping motions \cite{continuation, corosGeneralizedBipedWalking2010}. Spacetime optimization and reinforcement learning methods are then introduced to generate more complex motions \cite{GPS, CIO, yzqAthleticJumping}. These methods typically rely on carefully designed objective functions or hand-tuned parameters, and thus can be hard to apply to complex scenarios and general motion generation tasks.

In contrast, data-driven methods generate natural motions by imitating captured human performances in simulated environments \cite{liuImprovingSamplingbasedMotion2015a,liuGuidedLearningControl2016, fussellSuperTrackMotionTracking2021,pengDeepMimicExampleguidedDeep2018a,wangUniConUniversalNeural2020}. The use of motion capture data reduces the need to design specific objective functions for these methods. Instead, many of these approaches utilize a unified imitation error as their optimization target. By introducing perturbations and altering goals during training, the learned controllers can adapt to external disturbances, changes in target motions \cite{pengDeepMimicExampleguidedDeep2018a, xieSoccer2022}, and even variations in body shapes \cite{WonBodyShapeVariation2021, PerpetualHC}. Building on these successes, some works attempt to generate simulated motions by tracking the results of kinematic motion generators \cite{wonScalableApproachControl2020, wangUniConUniversalNeural2020, bergaminDReConDatadrivenResponsive2019, YuanyeTennis}. However, producing physically plausible animations using kinematic methods is challenging. The physical errors in the generated motions can degrade the performance of the tracking controller, making a native physics-based generative model more preferable.

\subsection{Generative Models in Physics-Based Motion Control}

In recent years, researchers have started to investigate the use of generative models within physics-based character animation, aiming to create diverse and flexible movements. Although studies in kinematic motion synthesis have shown the effectiveness of several generative models \cite{lingCharacterControllersUsing2020a,Li2022_GANimator,HenterMoGlow2020,Ao2023GestureDiffuCLIP}, this experience cannot be directly applied to physics-based motion control, because physics constraints are often considered as a black box and non-differentiable. A possible workaround is to learn motion primitives and reuse them in downstream tasks \cite{pengMCPLearningComposable2019, liuLearningScheduleControl2017,wonScalableApproachControl2020}. These motion primitives can also be distilled into a larger generative policy using behavior cloning \cite{merelCatchCarryReusable2020, merelNeuralProbabilisticMotor2018}, though the learned policy can suffer from distribution shift without feedback from actual simulation. More recent works draw inspiration from GAIL, or Generative Adversarial Imitation Learning \cite{HoGailNips2016}, and employ a discriminator as a motion prior to regulate the style of the character's motion \cite{pengAMPAdversarialMotion2021a, PengASELargescaleReusableAdversarial2022, xuGANLikeApproachPhysicsBased2021, PMP,compositeGAN, CALM}. This approach allows the character to complete complex tasks similar to example motions, such as interacting with scenes \cite{jason_character_scene}, objects \cite{PMP}, and other characters \cite{InteractionPrior}. However, the mode collapse issue makes it challenging to handle large-scale datasets \cite{PengASELargescaleReusableAdversarial2022}.

Our approach is built upon variational autoencoders (VAE)~\cite{Kingma2014_VAE}. The works most closely related to ours are \cite{ControlVAE, Won2022ConditionalVAE}, which also utilize a world model \cite{WorldModelHa2018} serving as differentiable simulator and reduce the task of physics-based motion generation to learning a variational autoencoder. However, both of these works capture only a few minutes of motion data. The selection of motion style in these methods is achieved by training another high-level policy on selected motion clips. In terms of integration with a natural language interface, \citet{PADL} also demonstrate several language-directed controllers for physics-based character animation. However, their datasets are also relatively small and necessitate a more meticulous segmentation into short motion clips. The language models in their approach are primarily employed to select the subsequent motion clip to execute, rather than facilitating generalized cross-modal generation. Unlike these previous works, our work is dedicated to enabling the model to learn diverse, task-agnostic motion encodings on a substantial unstructured dataset. Due to the abundance of motion encodings, we are able to learn a mapping from broader control tasks to physics-based motions.

\subsection{Motion Generation Tasks and Motion Representations}

Beyond the scope of physics-based character animation, our research is enriched by insights from the kinematic motion generation field,  particularly from areas like music-to-dance generation \cite{listen_denoise_act}, text-driven animations \cite{ACTOR,TEMOS,chen2023mld,tevet2023mdm}, and gesture synthesis \cite{Ao2023GestureDiffuCLIP,bailando}. Given the absence of physical constraints, these domains readily integrate with contemporary generative models, resulting in rapid advancements. 

The initial efforts are centered on learning a generative model directly in the motion space. These generative models include VAE \cite{ACTOR, TEMOS}, GAN \cite{Text2Motions} and flows \cite{gesture_flow, gesture_norm_ye}. Recently, \citet{tevet2023mdm, zhang2022lzw_motiondiffuse,physdiff} apply the diffusion model \cite{sohl2015diffusionmodel} to text-driven motion generation tasks and achieve notable success. Similarly, \citet{listen_denoise_act} utilizes the diffusion model to generate dance sequences and gestures, using audio and text as direct inputs. Due to the challenges of redundancy and noise in generating within the original motion space, many subsequent studies transition to a two-stage methodology, which first encode data into a latent space and afterwards learn a probabilistic model to generate the encodings \cite{bailando, chen2023mld, hml3d}. An optimal representation thus becomes essential for this line of research. Evidence from several recent studies indicates that \vqvae{} \cite{vqvae} provides an effective and compact motion representation \cite{rombach2022latentdiffusionimage, Geng23PCT, dhariwal2020jukebox}. As a result, many researchers opt to construct generative models within the \vqvae{} space \cite{dhariwal2020jukebox, DBLP:journals/tog/AoGLCL22,Ao2023GestureDiffuCLIP}. Additionally, leveraging the discrete nature of \vqvae{}, generative models from natural language processing (NLP), such as Generative Pretrained Transformer \cite{GPT3}, can be combined with motion models to achieve impressive results in cross-modal generation tasks \cite{bailando,t2mgpt,t2mt, jiang2023motiongpt}.

The advancements in kinematic motion generative models have demonstrated the benefits of training compact latent representations on large datasets and applying these representations to various downstream tasks. Inspired by these applications, especially those involving NLP models, we combined our motion representation with GPT, leading to physics-based text-to-motion generation, a downstream task that previous methods had not accomplished. Additionally, we showcase a practical integration with large language models.

\section{Discrete Motion Representation}
\label{sec:control_vq}

\begin{figure}[t]
    \centering
    \includegraphics[width=\linewidth]{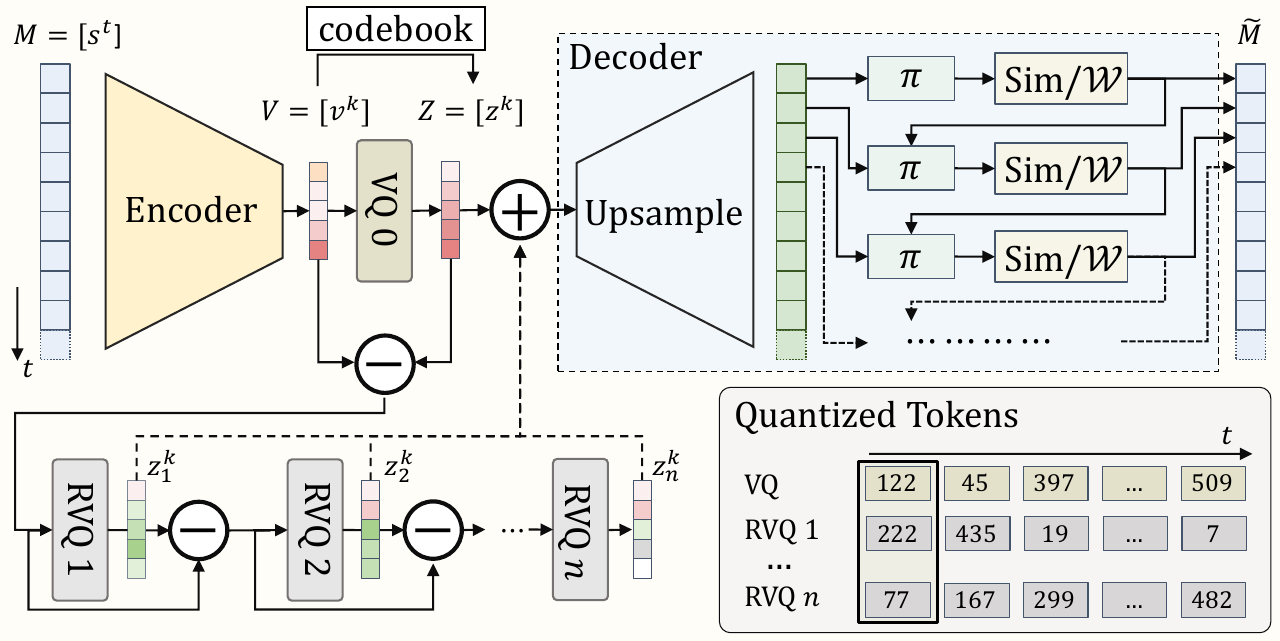}
    \caption{Framework overview. Our \frameworkname{} system consists of a motion encoder, a physics-based decoder, and a series of codebooks. A residual architecture is adopted to enhance the system's representational capacity. The system represents motion as a sequence of quantized latent codes. Each latent code combines the quantized vectors from all residual layers. Equivalently, it can be represented by the indices of these quantized vectors in the codebooks of their corresponding residual layers.}
    \label{Fig:Pipeline}
\end{figure}

\subsection{\frameworkname{} Framework}
\label{sec:rvq_encoder}

Our goal is to find a compact motion representation that captures the essential characteristics of a motion, enabling a character to reproduce it in a physics-based simulation environment. We utilize the Vector Quantized Variational Autoencoders (\vqvae{}) \cite{vqvae} to learn this representation from a large motion dataset encompassing a wide variety of motions. \fig\ref{Fig:Pipeline} illustrates the structure of our framework, which we refer to as \frameworkname{} in this paper.

Formally, a \frameworkname{} model consists of an encoder, $\encoder$, a decoder, $\decoder$, and a discrete codebook, represented as $\codebook$. The encoder $\encoder$ and the decoder $\decoder$ establish a bidirectional mapping between a motion clip, $\tarmo$, and its corresponding latent representation. The codebook $\codebook$ consists of a list of learnable latent codes $\{\latentcode_i\}_{i=1}^{\numcode},\latentcode_i \in \LatentSpace$, which is used to quantize the latent representation. We represent these procedures as
\begin{equation}
    \latentseq = \encoder (\tarmo), \quad 
    \latentvqseq = \codebook(\latentseq), \quad 
    \tilde{\tarmo} = \decoder( \latentvqseq ).
    \label{equ:encode_decode}
\end{equation}
The motion clip $\tarmo$ consists of a sequence of poses $[\stt^t]$. It is encoded by $\encoder$ into a sequence of latent vectors $\latentseq=[\latent^k]$. The code sequence $\latentvqseq=[\latentvq^k]$ is the quantized version of $\latentseq$. It is calculated by finding the nearest neighbor for each $\latent^k\in\latentseq$ within the codebook~$\codebook$:
\begin{equation}
    \latentvq^k, \vqidx^k = \arg \min_{\latentcode_i \in \codebook} \| \latentcode_i - \latent^k \|^2_2,
    \label{equ:normal_vq}
\end{equation}
where $\vqidx^k$ is the index of $\latentvq^k$ in $\codebook$. Notably, the quantized sequence $\latentvqseq$ can be represented using its corresponding indices, denoted as $\vqidxseq=[\vqidx^k]$. Finally, $\latentvqseq$, or equivalently, $\vqidxseq$, is decoded by $\decoder$ through a simulation process to reproduce the motion sequence $\tilde{\tarmo}$. 

We model the encoder $\encoder$ as a 1D convolutional neural network. It effectively downsamples a motion clip $\tarmo$ with  $T$ frames into a sequence of $K$ latent codes at a fixed ratio. The decoder, $\decoder$, on the other hand, is more complicated, comprising a 1D deconvolutional network, a control policy, and a simulation module. As shown in \fig\ref{Fig:Pipeline}, the deconvolutional network upsamples the quantized motion representation $\latentvqseq=[\latentvq^k]$ into a sequence of intermediate codes $[\midlatent^t] = \texttt{Deconv}([\latentvq^k])$,

which have the same framerate as the motion clip. Then, a policy, $\policy$, is trained to compute an action, $\act^t$, for each code $\midlatent^t$ based on the current state of the character, $\stt^t$. The action $\act^t$ is then used to actuate the character in the simulation, advancing the character to the next state, $\stt^{t+1}$. This control and simulation process can be written as
\begin{align}
    \act^t = \policy(\stt^t, \midlatent^t) ,\quad \stt^{t+1} = \texttt{Sim}(\stt^t, \act^t).
    \label{equ:real_sim}
\end{align}
It is performed autoregressively, using the simulated state from the previous step to compute the action and subsequent state. The resulting sequence of states, denoted as $[\stt^t]$, assembles the reconstructed motion $\tilde{\tarmo}$. This entire procedure, including the upsampling, control, and simulation, is collectively referred to as the decoder $\decoder$.

It is worth noting that the architectures of both the encoder and decoder allow them to process motions in a streaming manner. This means that, at runtime, the system can continuously feed new motion poses to the encoder. They are then encoded, quantized, and subsequently decoded into a sequence of simulated motion. Additionally, high-level policies for downstream tasks can generate and modify the latent codes in real-time, whether in the form of $[\latent^k]$, $[\latentvq^k]$, or $[\vqidx^k]$,  creating a continuous stream of simulated motions. We will discuss several example downstream tasks in Section~\ref{sec:applications}.

\subsection{Residual VQ-VAE} 
In practice, the representational capacity of a VQ-VAE model is restricted by the size of the codebook $\codebook$. In \frameworkname{}, as the volume and diversity of motion data increase, the potential variations within each small segment of data also expand. From an information theory perspective, to ensure that these nuances can be distinguished, more bits have to be allocated to represent them. This leads to an exponential growth in the size of the codebook, resulting in inefficiency and unstable training due to code collapse \cite{dhariwal2020jukebox}. Similar facts are noted in works related to audio compression \cite{zeghidour2021soundstream} and image generation \cite{lee2022autoregressive}.

Residual \vqvae{} \cite{zeghidour2021soundstream,lee2022autoregressive} addresses such issues by utilizing a residual architecture with multiple quantization layers. As depicted in \fig\ref{Fig:Pipeline}, in addition to the standard vector quantization of \vqvae{}, the residual \vqvae{} employs a series of additional residual vector quantization (RVQ) layers and corresponding learnable codebooks $\{\codebook_d\}_{d=1}^{N}$ to model the residues. Specifically, the first RVQ layer (RVQ-1) considers the residue resulting from the \vqvae{}'s quantization, given by ${r}_1 = \latent - \latentvq$. It then looks up the codebook $\codebook_1$ and quantizes ${r}_1$ into $\latentvq_1$. For a subsequent RVQ layer, RVQ-$d$, the residue of the previous RVQ layers, defined as ${r}_d = \latent - \sum_{j<d} \latentvq_j$, is quantized into $\latentvq_d$ based on the corresponding codebook $\codebook_d$. If we regard the standard \vqvae{} as a special initial RVQ layer (RVQ-0), this entire process can be viewed uniformly, as outlined in Algorithm~\ref{alg:rVQ}. With the RVQ layers, the quantized code of each $\latent^k \in \latentseq$ is the sum of all RVQ codes, \ie{}, $\latentvq^k=\sum_{d=0}^N \latentvq_d^k$, or equivalently, the tuple of indices of residual codes within each RVQ layer, $\vqidx^k=(\vqidx_0^k,\vqidx_1^k,\dots,\vqidx_N^k)$, where $\vqidx_d^k$ is the index of $\latentvq_d^k$ in the codebook $\codebook_d$.

\begin{algorithm}[t] 
    \DontPrintSemicolon
    \SetAlgoLined
    \SetKwProg{RVQ}{Function}{:}{end}
    \RVQ{ \textnormal{\textbf{Residual Vector Quantization}$(\latent)$} }
    {
        \tcp{$\latent\in\latentseq$ is the continuous latent code}
        \tcp{computed using \eqn\eqref{equ:encode_decode}}
        $\latentvq \leftarrow 0$\;
        $r \leftarrow \latent$ \;
        
        \For{$d\gets0$ \KwTo $N$ }{
            $\latentvq_d, \vqidx_d
            \leftarrow 
            \argmin_{\latentcode_i \in \codebook_d} \|{r} - \latentcode_i  \|_2^2 $\;
            
            ${r} \, \mathrel{-}= \latentvq_d$ \;
            $\latentvq \, \mathrel{+}= \latentvq_d$\;
        }
        $\vqidx \gets (\vqidx_0, \vqidx_1, \dots, \vqidx_N)$ \;
        \Return{$\latentvq$, $\vqidx$}
    }
  
  \caption{Algorithm of Residual Vector Quantization}
  \label{alg:rVQ}
\end{algorithm}

Compared to directly increasing the size of the codebook, the residual \vqvae{} expands the capacity of the model exponentially as the number of layers increases \cite{zeghidour2021soundstream}. Another benefit is that it establishes a coarse-to-fine representation of the entire latent space. Here, the coarse latents capture the primary components of the motion, while the fine-grained latents enrich it with details. For a more detailed discussion, please refer to Section~\ref{sec:ablation}.

\subsection{Model-Based Learning of \frameworkname{}}
\label{sec:pdecoder}

We train our \frameworkname{} model on a large and diverse dataset comprising several tens of hours of motion data. An efficient and stable training algorithm is crucial to successfully handle such a volume of data. Additionally, we treat the simulation as a black box. While this allows our framework to be independent of specific physics engines, it also makes the decoder non-differentiable. This prevents the use of the standard training method for \vqvae{}, which requires the gradient to pass through the entire model.

Inspired by recent works that employ model-based reinforcement learning for complex motion controllers \cite{fussellSuperTrackMotionTracking2021,ControlVAE}, we adopt a model-based method to train our \frameworkname{} models. Specifically, we jointly train a network, referred to as the world model $\worldmodel$, to approximate the black-box simulator. The control and simulation process described by \eqn\eqref{equ:real_sim} then becomes
\begin{align}
    \hat{\act}^t = \policy(\hat{\stt}^t, \midlatent^t) ,\quad \hat{\stt}^{t+1} = \worldmodel(\hat{\stt}^t, \hat{\act}^t),
    \label{equ:world_model_sim}
\end{align}
where the simulation operator \texttt{Sim} is replaced by the world model $\worldmodel$. This adaptation ensures that the decoding of the reconstructed motion sequence $\tilde{\tarmo}_{\worldmodel}=[\hat{\stt}^t]$ is perfectly differentiable. We use $\decoder_{\worldmodel}$ to denote this decoding process and a hat ($\hat{\ \ }$) to indicates a quantity synthesized by the world model.

We train the world model and \frameworkname{} using an approach similar to  ControlVAE~\cite{ControlVAE}. In each training iteration, the world model and the components of the \frameworkname{} model are alternately updated until convergence. For the world model $\worldmodel$, we first instruct the character to track a random reference motion $\tarmo$ using the current \frameworkname{}. This process involves encoding $\tarmo$ into latent codes, quantizing them, and decoding the quantized codes in real simulation, resulting in a simulated motion sequence $\tilde{\tarmo}_{\text{sim}}$. Then, we replay $\tilde{\tarmo}_{\text{sim}}$ with $\decoder_{\worldmodel}$ and $\worldmodel$ to generate a synthesized sequence $\tilde{\tarmo}_{\worldmodel}$. After this, the world model $\worldmodel$ is updated by optimizing the loss function
\begin{equation}
    \loss_{\worldmodel} = \| \tilde{\tarmo}_{\text{sim}}  - \tilde{\tarmo}_{\worldmodel} \| .
    \label{equ:world_model_loss}
\end{equation}
Subsequently, the components of the \frameworkname{} model, including the encoder $\encoder$, the codebook $\codebook$, and the policy $\policy$, are updated simultaneously. The character tracks a new training motion $\tarmo$ with the world model $\worldmodel$, creating a new synthesized reconstruction $\tilde{\tarmo}_{\worldmodel}$. Then, the \frameworkname{} model is updated by optimizing the loss function
\begin{align}
    \mathcal{L} &= \| \tarmo - \tilde{\tarmo}_{\worldmodel} \| \nonumber\\
    &+ \beta_1 \| \encoder(\tarmo) - \texttt{sg}(\latentvqseq) \nonumber\| 
      + \beta_2 \| \texttt{sg}(\encoder(\tarmo)) - \latentvqseq \nonumber\| \\
    &+ \beta_3 \mathcal{L}_{\text{reg}} \quad, 
    \label{equ:vq_vae_loss}
\end{align}
where the first three terms correspond to the standard VQ-VAE losses \cite{vqvae}, namely the reconstruction loss, the commitment loss, and the VQ objective. \texttt{sg} represents the stop gradient operator. To prevent the policy from generating excessively large output, which could lead to unnatural motions and cause unstable training, we introduce a regularization term, $\mathcal{L}_{\text{reg}}$, for the actions. Further details are discussed in the next section.

\subsection{Implementation}

\paragraph{State} Our simulated character is modeled as an articulated rigid body system with a floating base. Similar to the frameworks presented in \cite{fussellSuperTrackMotionTracking2021,ControlVAE}, we define the state of the character, $\stt$, using the following attributes:
\begin{itemize}
    \item Rotation ${r}$ and position ${p}$ of each body
    \item Velocity ${v}$ and angular velocity ${\omega}$ of each body
    \item Height $h$ of each body
    \item Up direction ${y}_0$ of the root joint
\end{itemize}
The root joint is designated to be in the character’s pelvis. The attributes ${r}, {p}, {v}, {\omega}$ are represented in the local coordinate frame of the root joint.

\paragraph{Encoder and Decoder} We formulate the encoder $\encoder{}$ of \frameworkname{} as a 1D convolutional network. Together with the upsampling module \texttt{Deconv} of the decoder $\decoder$, which is also a 1D convolutional network, they form a symmetric fully convolutional network. The design and structure of this combined network are similar to those in \cite{dhariwal2020jukebox} and \cite{Ao2023GestureDiffuCLIP}, which focus on generative models for time-series data such as audio \cite{dhariwal2020jukebox} and gesture \cite{Ao2023GestureDiffuCLIP}.

\paragraph{Policy and Action}
The policy $\policy$ of the decoder $\decoder$ adopts a layer-wise mixture of experts (MoE) model as introduced by \citet{xieSoccer2022}. This model consists of six experts, each being a four-layer MLP with 256 units, and their outcomes are mixed based on blending weights determined by a gating network, which is a two-layer MLP with 64 units. 

We actuate the character using joint torques. Each joint torque is computed using the widely-adopted PD control scheme
\begin{align}
    \tau = k_p(\bar{\theta} - \theta) - k_d \dot{\theta},
\end{align}
where $\theta$ and $\dot{\theta}$ denote the current joint rotation and rotational speed, respectively. $\bar{\theta}$ represents the desired joint rotation. $k_p$ and $k_d$ are the proportional and derivative gains, respectively. The output of the policy $\policy$, \ie{}, the action vector $\act$, thus comprises the desired joint rotations of all the joints.

\paragraph{Action Regularization.} While the quantization of \frameworkname{} offers robustness to minor perturbations in training motions, flaws and physically implausible movements in the training motion data can degrade the performance of a \frameworkname{} model. For instance, the model might misinterpret high-frequency noise in the input motion as specific motion details, leading to oscillating movements. Similarly, for motions captured on a treadmill, the model may abuse the imperfect contact model of the physics engine and create foot-skating to mimic the stationary motion.

We notice similar phenomena have been reported by previous research, such as \cite{bergaminDReConDatadrivenResponsive2019} and \cite{spring_mass}. 
\citet{bergaminDReConDatadrivenResponsive2019} use exponential moving average (EMA) with a large smoothing factor to mitigate this issue. However, in practice, we find that such a hard constraint can overly smooth the motions, eliminating many subtle motion details. Instead, we employ EMA as a soft constraint in the optimization problem of \eqn\eqref{equ:vq_vae_loss}, which becomes the regularization loss $\loss_{\text{reg}}$. Specifically, when creating the synthesized motion sequence $\hat{\tarmo}$, we compute the EMA of the actions as
\begin{equation}
    \bar{\act}^t = (1-\beta) \bar{\act}^{t-1} + \beta \hat{\act}^t ,
\end{equation}
where $\beta$ represents the smoothing factor. Empirically, we find that $\beta=0.8$ achieves reasonable visual results. $\loss_{\text{reg}}$ is then computed as
\begin{equation}
    \loss_{\text{reg}} = \sum_{t} w_1\| \hat{\act}^t - \bar{\act}_t \|+ w_2\|\hat{\act}^t\|
\end{equation}
where the first term limits the difference between the action computed by the policy $\policy$ and its EMA, encouraging a smooth sequence of actions. The second term regularizes the absolute magnitude of the action to enforce training stability, which is also used in \cite{fussellSuperTrackMotionTracking2021, ControlVAE}.

\paragraph{World Model} The differentiable world model $\worldmodel$ consists of a four-layer MLP with $512$ units, similar to those used in previous research \cite{fussellSuperTrackMotionTracking2021, ControlVAE, Won2022ConditionalVAE}. In practice, we employ a large first-in-first-out replay buffer to store the simulated motion sequences $\tilde{\tarmo}_{\text{sim}}$. We then evaluate the loss function \eqn\eqref{equ:world_model_loss} using a batch of sequences sampled from this buffer. This approach, suggested by \cite{fussellSuperTrackMotionTracking2021, ControlVAE}, significantly stabilizes training and is critical to its success. Our replay buffer has a size of $50000$, and we replace $1024$ simulated motion frames in each iteration.

\paragraph{Training Techniques for Residue \vqvae{}} In addition to the training procedure discussed in the previous section, we also employ several techniques proposed in previous works \cite{vqvae, t2mgpt, zeghidour2021soundstream, jiang2023motiongpt, DBLP:journals/tog/AoGLCL22, bailando} to ensure a successful training. These techniques include 
\begin{itemize}
    \item \textbf{EMA Update}: Updating the codebook $\codebook$ using an exponential moving average scheme \cite{ t2mgpt, zeghidour2021soundstream, jiang2023motiongpt, bailando} formulated as 
    \begin{equation}        
        \codebook \gets (1-\gamma) \codebook + \gamma \hat{\codebook} ,
    \end{equation}
    where $\hat{\codebook}$ represents the optimized codebook and $\gamma$ is the smoothing factor. 
    \item \textbf{Code Reset}: Resetting codes that are inactive during training \cite{t2mgpt, Ao2023GestureDiffuCLIP,dhariwal2020jukebox}. This can alleviate the codebook collapse problem in \vqvae{} training. 
    \item \textbf{Quantizer Dropout}: Using a random span of the first RVQ layers, instead of using all layers, to reconstruct the motion during training \cite{zeghidour2021soundstream}. The number of these RVQ layers is selected randomly from $\left[1, 2, \dots, N\right]$ in each iteration. This approach ensures that motions can be represented using any span of the initial RVQ layers, allowing a tunable balance between representation complexity and motion reconstruction quality without needing to retrain the system. 
\end{itemize}
\section{Applications and Tasks}
\label{sec:applications}

A trained \frameworkname{} model provides a foundation that enables various downstream tasks to be tackled within a unified framework. We categorize these tasks into two groups: \emph{encoder-decoder tasks} and \emph{decoder-only tasks}. For the \emph{encoder-decoder tasks}, the latent motion representation is computed by the pretrained encoder, which processes motion sequences generated by other methods. We also refer to this type of task as \emph{tracking control}. For the \emph{decoder-only tasks}, we train a separate high-level task policy to compute the sequence of latent codes based on specific conditions. These latent codes are then decoded into motions through simulation. This \emph{decoder-only} configuration also allows for direct integration with a large language model (LLM), enabling more complex tasks that can leverage the reasoning capabilities of the LLM. In this section, we will explore several representative applications and tasks to showcase the potential of this framework.

\subsection{Universal Tracking Control}

We first demonstrate the capacity of our \frameworkname{} framework in tracking problems, using the \emph{encoder-decoder} configuration. Particularly, an input motion is encoded using $\encoder$, quantized according to $\codebook$, and decoded with $\decoder$ through simulation. The quantized embedding effectively eliminates disturbances in the original motion, enables imitation of unseen motions or even noisy inputs, and allows the creation of physically correct motions. We highlight the tracking capabilities of our \frameworkname{} by testing it on motions from three different sources, including:
\begin{itemize}
    \item \textbf{Unseen motion}: Motion clips from a motion dataset not used in training.
    \item \textbf{Kinematic motion generation}: Motion clips generated by a motion synthesis method without involving simulation. Such motions can sometimes contain artifacts like foot-skating, self-penetration, or penetrating other objects.
    \item \textbf{Video-based pose estimation}: Motion clips estimated from a video using a pose estimation approach. These motions can often be noisy and lack physical accuracy.
\end{itemize}
We show results and analysis in Section \ref{sec:tracking_results}.

\subsection{Interactive Control}

We further demonstrate that our \frameworkname{} framework is capable of incorporating a high-level policy to achieve interactive control using the \emph{decoder-only} configuration. In this setup, the high-level policy computes latent codes directly based on the current state and task requirements. While reinforcement learning is frequently used to train such hierarchical control policy, we show that the robustness offered by the quantized motion representation enables even more efficient supervised learning.

We take \emph{steering control} as an example, where the character is required to respond to user input and move in the desired direction and speed. We employ a simple autoregressive model as the control policy, similar to previous work such as \cite{DanielPFNNPhaseFunctionedNeuralNetworks2017}. This policy, $\policy_{\text{steering}}$, is formulated as 
\begin{align}
    \latent^k, {\task}^{k+1}_{\pi} = \policy_{\text{steering}}(\latentvq^{k-1}, \task^k),  \quad \latentvq^k = \codebook(\latent^k)
    \label{equ:steering_policy}
\end{align}
where $\task^k$ denotes the task parameters, which is computed according to the difference between the character's current state and the desired direction and speed of movement. $\latentvq^{k-1}$ represents the quantized latent code computed from the previous control time step. The output of the policy is a new latent vector $\latent^k \in \LatentSpace$, which is then quantized as described in Section~\ref{sec:control_vq}. Following \cite{DanielPFNNPhaseFunctionedNeuralNetworks2017}, we also let the policy predict the future task parameters, denoted by ${\task}^{k+1}_{\pi}$, which is used to facilitate training and improve policy performance during inference.

We train this policy by distilling the strategy embedded in a reference dataset consists of several locomotion sequences. During the training stage, we encode the entire dataset into quantized latent codes using the pretrained encoder $\encoder$, and compute the corresponding task parameters. This training dataset can then be represented as $[({\latentvq}^k_*, {\task}^k_*)]$. Subsequently, by unrolling \eqn\eqref{equ:steering_policy}, we generate a sequence of predictions, $[(\latent^k, {\task}^{k+1}_{\pi})]_{k=0}^K$ and $[\latentvq^k]_{k=0}^K$. Then, the policy is updated by optimizing the loss function
\begin{equation}
    \loss_{\policy} = \sum_{k=0}^{K} \| ({\latentvq}^{k}_*,{\task}^{k+1}_*) - \policy_{\text{steering}}(\latentvq^{k-1}, {\task}^{k}_*) \|.
\end{equation}

To encourage the network to respond to control signals, we jointly train a projector network $P$ which takes in the synthesized sequence $[\latentvq^k]_{k=0}^K$ and predicts the initial control signal $\task^0_*$. The final training loss is formulated as
\begin{equation}
    \loss_{\text{steering}} = w_1 \loss_{\policy} + w_2\left\|P\left([\latentvq^k]_{k=0}^K\right) - \task^0_*\right\|
\end{equation}

Following \cite{lingCharacterControllersUsing2020a}, we employ a scheduled sampling mechanism during the training process. Specifically, when unrolling \eqn\eqref{equ:steering_policy}, the latent code input to the policy, $\latentvq^{k-1}$, is randomly selected between the generated code and the reference latent code $\latentvq^{k-1}_*$. The probability of selecting the reference code $\latentvq^{k-1}_*$ is $p$, while the probability of choosing the generated code $\latentvq^{k-1}$ is $1-p$. We initially set the probability $p$ to $0.8$ at the beginning of the training, and gradually decrease it to $0$.

The task parameters $\task^k$ comprise the position and facing direction of the character over the upcoming $0.9$ seconds, sampled every $0.3$ seconds. During training, these properties are extracted from the reference motions and are computed relative to the frame $k$. At runtime, we compute a set of desired task parameters $\task_u$ based on user input, in a manner similar to the approach in \cite{DanielPFNNPhaseFunctionedNeuralNetworks2017}. Subsequently, $\task_u$ is blended with the task parameters ${\task_{\pi}}$ predicted by the policy. The final parameters ${\task}$ input to the policy are computed as

\begin{equation}
    \task = (1-\tau^{\gamma})\odot {\task_{\pi}} + \tau^{\gamma} \odot  \task_u
    \label{equ:blend}
\end{equation}
where $\tau$ is a vector of blending weights. We use $\tau=[1/3, 2/3, 1]$ for the three sampled frames for computing $\task$. The operator $\odot$ denotes the element-wise multiplication between two arrays. The parameter $\gamma$ is a factor to adjust the blending speed and is set to $0.5$ for position and $2$ for rotation, as suggested in \cite{DanielPFNNPhaseFunctionedNeuralNetworks2017}.

\subsection{Generative Motion Prior with GPT}

A generative motion prior models the probability distribution of human motions, $p(\tarmo)$, or equivalently, $p(\stt^t|\stt^{<t})$. Recent works in generative motion control indicate that this distribution, especially its simplified version $p(\stt^t|\stt^{t-1})$, can be effectively modeled using a simple forward network \cite{pengAMPAdversarialMotion2021a, PengASELargescaleReusableAdversarial2022,Won2022ConditionalVAE} or a mixture-of-expert structure \cite{xieSoccer2022,ControlVAE} when trained on specific motions with a small amount of example data.  However, for a broader and more diverse set of motions, there is still a need for a more comprehensive model to capture the intricate dynamics and subtle variations within the motions. 
One popular and effective approach in recent research is to integrate \vqvae{} with the Generative Pretrained Transformer (GPT) \cite{GPT3}, which has gained great success in natural language processing \cite{chatgpt} and image generation tasks \cite{AnImageIsWorth16words} and has also found applications in kinematic motion generation \cite{t2mgpt, jiang2023motiongpt}. 

In this task, we demonstrate that a physics-based motion GPT, which we term \emph{\controlGPT{}}, can be built upon our \frameworkname{} representation and can serve as an efficient generative motion prior. This model is also based on the \emph{decoder-only} configuration. Rather than using the latent codes, we adopt the index-based motion representation for this task. We refer to this approach as the \emph{index-decoder-only} configuration.

Formally, given a motion sequence, $\vqidxseq = [\vqidx^k]$, represented as a sequence of indices corresponding to quantized latent codes, 
we train a transformer-based network \cite{AttentionIsAllYouNeed2017} to predict each motion index given the previous indices from the same motion sequence. This prediction corresponds to the conditional probability distribution $p(\vqidx^k|\vqidx^{<k})$. The probability of the motion sequence $\vqidxseq$ can then be computed as
\begin{align}
    p(\vqidxseq) = p(\vqidx^0) \prod_{k=1}^{K} p(\vqidx^k|\vqidx^{<k}),
\end{align}
where $K$ is the length of $\vqidxseq$.

\begin{figure}[t]
    \centering
    \includegraphics[width=\linewidth]{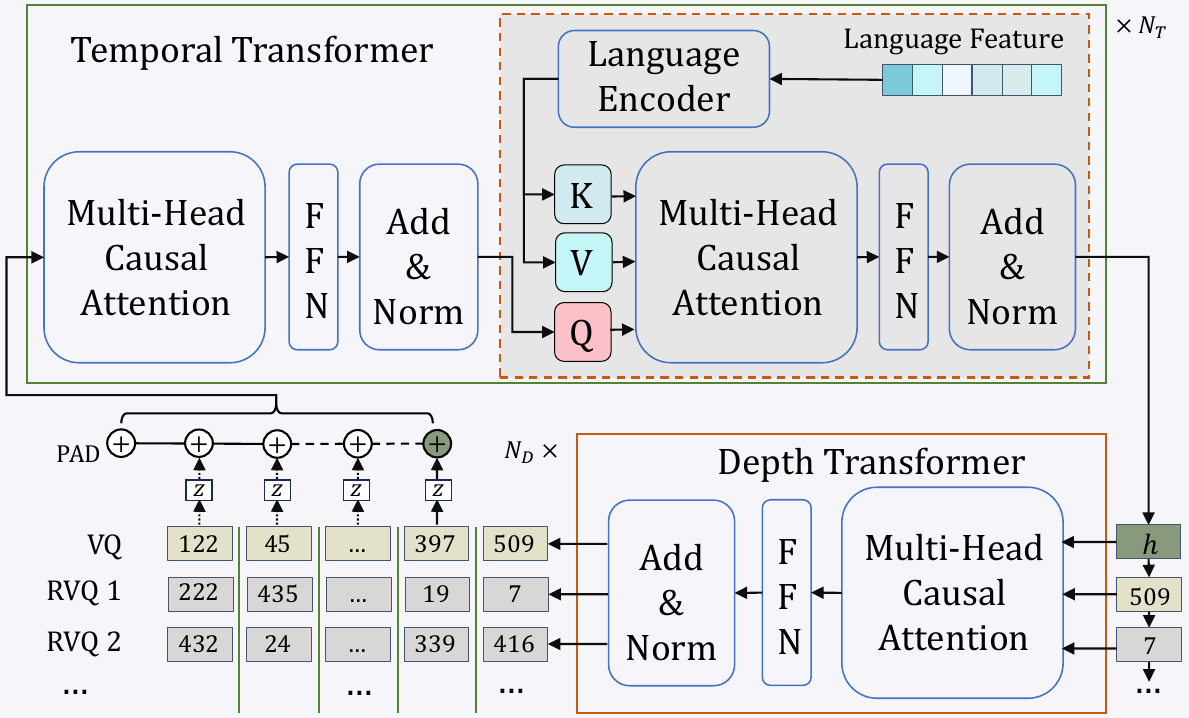}
    \caption{Architecture of (T2M-)\controlGPT{}. We adopt a dual-transformer architecture to predict the next motion code in the form of RVQ indices given the previous codes. This architecture can be extended to accept additional features to generate motions based on descriptions in natural language.}
    \Description{}
    \label{Fig:transformer}
\end{figure}

Considering that the RVQ mechanism is adopted, each $\vqidx^k$ represents a tuple of RVQ indices $\vqidx^k=[\vqidx^k_d]_{d=0}^{N}$. While we could technically flatten the RVQ indices of all motion codes of $\vqidxseq$ into a very long sequence and apply GPT using a single large transformer, such a naive architecture does not fully exploit the hierarchical structure of the residual \vqvae{}. Moreover, it would be computationally demanding since the computational cost of a transformer increases with the square of the sequence length it needs to consider. As an alternative, we take inspiration from \cite{lee2022autoregressive} and model the dynamics of the index sequence using two separate transformers. Particularly, as depicted in \fig \ref{Fig:transformer}, we compute
\begin{align}
    p(\vqidx^k|\vqidx^{<k}) =  \prod_{d=0}^{N} \underbrace{p\bigl(\vqidx^k_d \,\big|\,  \overbrace{  f(\vqidx^{<k})  }^{\clap{$\substack{\text{temporal} \\ \text{transformer}}$}} \,,\, \vqidx^k_{<d} \bigr)}_{{\text{depth transformer}}}, 
    \label{equ:control_gpt}
\end{align}
where the \emph{temporal transformer} aggregates information from previous motion codes, and the \emph{depth transformer} calculates RVQ indices based on both the aggregated features and the indices from preceding RVQ layers. To achieve a compact network architecture, we convert each $\vqidx^{<k}$ into the corresponding quantized latent code $\latentvq^{<k}$ before inputting it into the \emph{temporal transformer}.

We train the motion prior on a large dataset that consists of many motion sequences. From these sequences, we extract the discrete index-based representations using the pretrained \frameworkname{} encoder. We then optimize the model using a negative log-likelihood (NLL) loss:
\begin{equation}
    \mathcal{L}_{\text{GPT}} = \mathbb{E}_{\vqidxseq}  \left[\sum_{k=1}^{K} \sum_{d=0}^N  - \log p \bigl( \vqidx^k_d \,\big|\, f(\vqidx^{<k}), \vqidx^k_{<d} \bigr) \right] ,
\end{equation}
where $\vqidxseq$ is randomly selected from the dataset during training.

\paragraph{Unconditional Motion Generation} 
Starting from a random initial motion index $\vqidx^0=[\vqidx^0_d]$, we can repeatedly sample the next motion index from the distribution in \eqn\eqref{equ:control_gpt} using a learned \controlGPT{} model. This generated sequence of motion indices is then streamed into the decoder of \frameworkname{}, converting it into a motion sequence. Such unconditional motion generation can produce random high-quality motions. The physics-based control automatically ensures a natural response to unexpected environmental changes and perturbations.

\paragraph{Text2Motion Generation}
The aforementioned unconditional \controlGPT{} model can be augmented to accept conditions such as text description. As illustrated in \fig\ref{Fig:transformer}, we integrate text features into our temporal transformer using an additional cross-attention layer. This conditional \controlGPT{} model is trained on a dataset containing paired motion and text description data. We process the motion as described above and utilize a pretrained T5 encoder \cite{raffel2020exploringt5} to encode the text data. This model, which we term \emph{T2M-\controlGPT{}}, effectively enables the generation of high-quality motions based on descriptions in natural language.

\subsection{Integration with Large Language Models}

Large Language Models (LLMs), such as ChatGPT \cite{chatgpt}, have been shown not only to be effective language models for NLP tasks but also potentially powerful tools for problem-solving \cite{shinn2023reflexion, yao2023react}, task planning \cite{COT, TOT}, and even acting as autonomous agents \cite{Park2023GenerativeAgents}. 

In this section, we shift our focus toward leveraging pretrained LLMs to equip our character with higher motion intelligence. We employ the \emph{index-decoder-only} configuration, but only consider the VQ index, omitting the RVQ layers for simplicity. We use the framework commonly referred to as in-context learning (ICL) for this task. In ICL, the pretrained LLM is provided with several example queries and answers, from which it learns to respond to new queries. Using this paradigm, we feed the LLM several thousand lines of text. Each line contains a text description and an index sequence extracted from the corresponding motion using the pretrained \frameworkname{} encoder. We then query the LLM with various questions and instruct it to respond with sequences of numbers. These numbers are interpreted as motion index sequences and are decoded by the physics-based decoder of \frameworkname{}. Surprisingly, without any fine-tuning, the LLM can readily adapt to \frameworkname{}'s index-based motion representation in such in-context learning. With the modest dataset of examples, the LLM demonstrates the ability to understand the meaning encoded within the example motion index sequences. It can then reconstruct motion phrases and skillfully recombine them to generate new motions.

\begin{figure}[t]
    \centering
    \includegraphics[width=\linewidth]{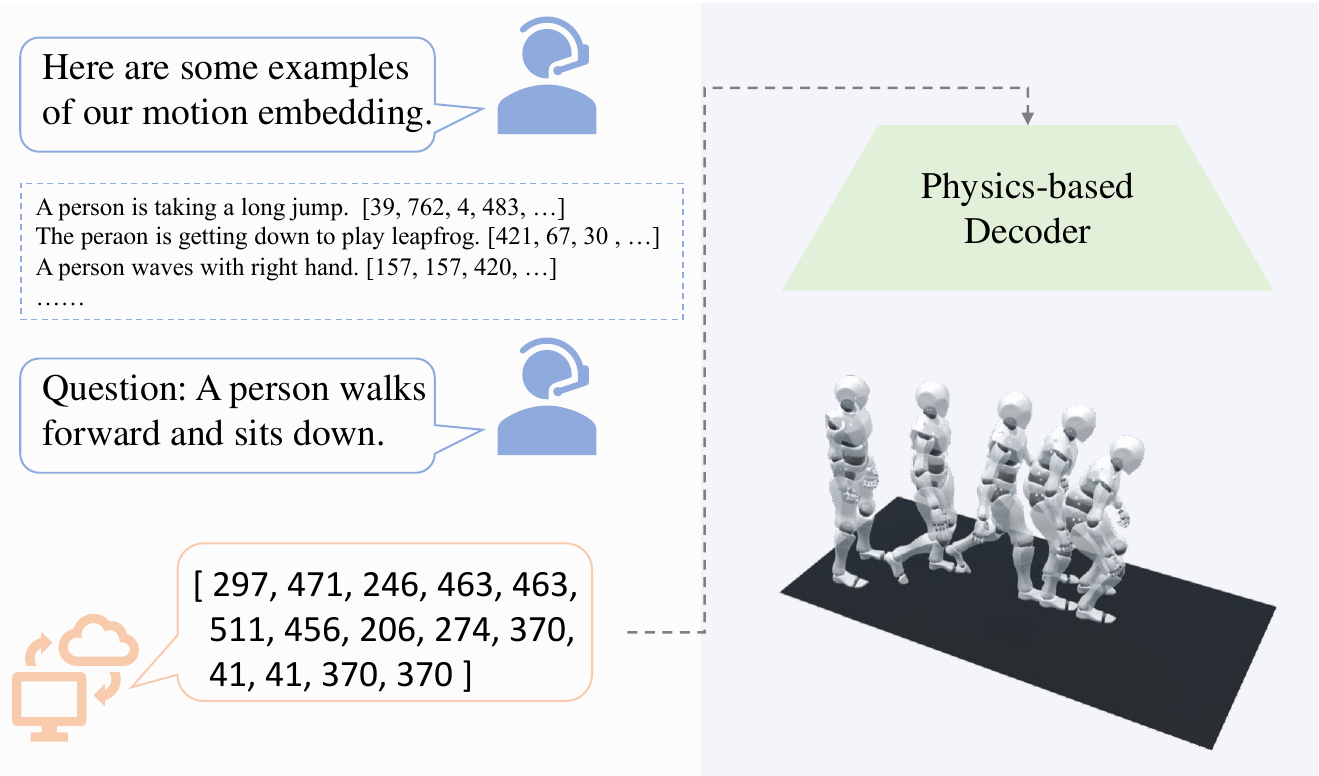}
    \caption{Integration with Large Language Models (LLMs). We can instruct a pretrained LLM, such as ChatGPT and Claude-2, to learn to utilize the motions embedded in \frameworkname{}. The latent motion indices serve as a compact representation that enables such integration via in-context learning.}
    \label{Fig:ChatGPT}
\end{figure}

Figure \ref{Fig:ChatGPT} illustrates an example of integration with a commercial LLM, Claude-2 \cite{claude}. After providing the examples, we input the following query: \textPrompt{question: a person walks forward and sits down}. In response, the LLM generates a sequence of indices in the form of: 
\begin{itemize}
    \item \textPrompt{297, 471, 246, 463, 463 - Person walks forward}
    \item \textPrompt{511, 456 - Person bends knees to sit down}
    \item \textPrompt{206, 274, 370 - Person sits on the ground}
    \item \textPrompt{41, 41, 370, 370 - Person stands back up from sitting}
\end{itemize}
By decoding this sequence of indices, \textPrompt{[297, 471, 246, 463, 463, 511, 456, 206, 274, 370, 41, 41, 370, 370]}, the character walks and then imitates the sitting motion in the simulation. Notably, the query description is neither contained in the examples we provided, nor does the particular combination of actions exist in the training data of the \frameworkname{} model. Instead, the LLM learns to decipher the meaning of the example index clips and autonomously recombines them to address the question.

This task highlights an important capability of our \frameworkname{} framework: in addition to merely translating text into motion, the \frameworkname{} embeddings can serve as a bridge between an LLM and the character's motor skills. Intuitively, one can analogize the LLM as the \emph{brain} of the character, driving its intelligence and comprehension, while the \frameworkname{} can be thought of as its \emph{cerebellum}, managing coordination and motor skills. Once the LLM grasps the skills embedded in the \frameworkname{} model, it can leverage this knowledge to direct the character to undertake more complicated and abstract tasks in a simulated environment. An in-depth exploration of this is provided in Section~\ref{sec:chatgpt}.

\section{Experiments and Results}

\subsection{System Setup}

\paragraph{Character}

Our simulated character is modeled as a floating-based articulated rigid body system, comprising $19$ joints and $20$ bodies. It has a height 1.6\,m and weighs 49.5\,kg. The character is simulated using a customized Open Dynamics Engine (ODE) \cite{OpenDynamicsEngine}, where an implicit damping scheme \cite{Liu2013_Simulation,tanStableProportionalDerivativeControllers2011b} is implemented to allow stable simulation at 120\,Hz. The PD control paramters are unformally set to $(k_p, k_d)=(400, 50)$ for all joints, except the toe joints $(10,1)$ and wrist joints $(5,1)$.  The control policy of the \frameworkname{}'s decoder is executed at 20\,Hz, where the computed PD target is used for the subsequent simulation steps until the next evaluation of the control policy.

\paragraph{Dataset} 
We train our \frameworkname{} system on a motion dataset consisting of 23.2 hours of a diverse range of motions, from basic locomotion to dynamic stunts such as dancing, boxing, and kung fu actions. These motions come from the LaFAN dataset \cite{harvey2020robustLafanYHY} and the AMASS dataset \cite{AMASS:ICCV:2019}. For the LaFAN dataset, we include all data except those involving interactions with objects. As for the AMASS dataset, which encompasses many individual motion capture datasets, we selectively populate our dataset with subsets that are both extensive in scale and of high quality. Before training, this dataset is donwsampled to 20\,FPS and is retargeted to our simulated character by copying the rotations of the corresponding joints. Statistical data of the training dataset are listed in Table~\ref{tab:motion_dataset}.

\begin{table}[t]
    \centering
    \caption{Motion dataset. LaFAN is from \cite{harvey2020robustLafanYHY}, the other datasets are from AMASS \cite{AMASS:ICCV:2019}}
    \label{tab:motion_dataset}
    \begin{tabular}{ccl}
        \toprule
        Dataset & Length (min) & Motions\\
        \midrule
        SFU     & 10.90  & Balance, Dance, Kendo, Skip, ... \\
        ACCAD   & 14.61  & Run, Swagger, Punch, Turn, ... \\
        BMLmovi & 103.32 & Walk, Stretch, Throw, Catch, ... \\
        BMLrub  & 180.52 & Lift, Sit, Kick, Throw, ... \\
        CMU     & 375.24 & Sports, Dance, Pantomime, ... \\
        KIT     & 392.29 & Wipe, Bend, Parkour, Drink, ... \\
        LaFAN   & 213.14 & Walk, Dance, Aiming, Fight, ... \\
        \bottomrule
    \end{tabular}
\end{table}

\begin{figure}[t]
  \centering
  \includegraphics[width=\linewidth]{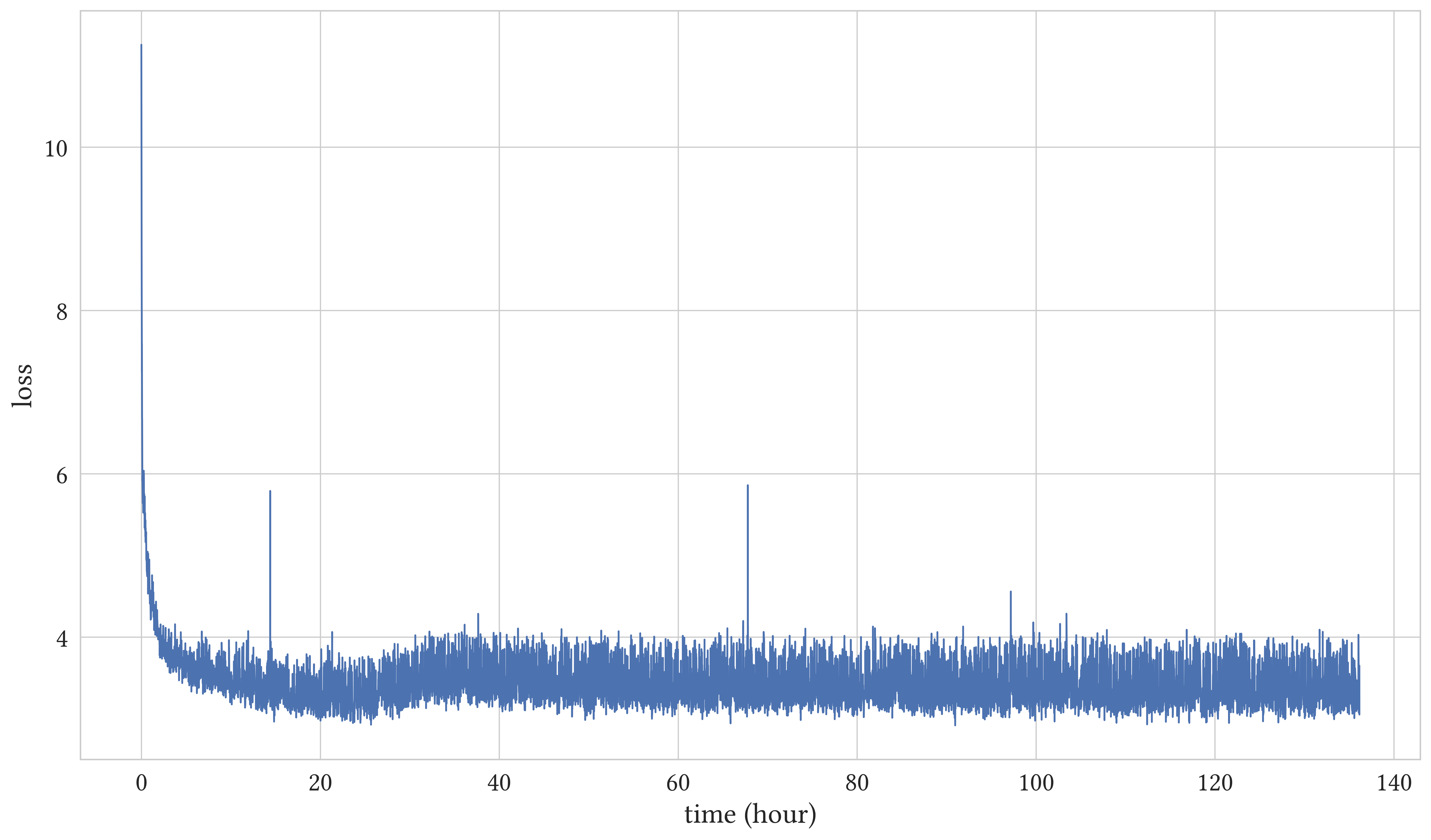}
  \caption{A typical learning curve of \frameworkname{}. }
  \label{fig:LossCurve}
\end{figure}

\paragraph{Settings} 
We train our \frameworkname{} on a single NVIDIA RTX 3090 GPU and five cores of an Intel(R) Xeon(R) E5-2678 v3 @ 2.50GHz CPU. \fig\ref{fig:LossCurve} presents a typical learning curve. For training, we use a fixed motion length of $T=24$ frames, equivalent to $1.2$ seconds. Each motion clip is encoded into $K=6$ latent codes, with each code having a size set to $768$. In our framework, we employ $N=8$ RVQ layers. All the codebooks contain $|\mathcal{B}|=512$ code vectors. We set the smoothing factor of the EMA for updating the codebooks to $\gamma = 0.99$. For optimization, we use the RAdam optimizer \cite{liu2019radam}, with a learning rate of $1\times 10^{-5}$. The model is trained for 40k epochs, taking about six days. The motion quality can see further improvements as training continues.

\subsection{Universal Tracking Control}

\label{sec:tracking_results}
We evaluate the performance of our \frameworkname{} model in the \emph{encoder-decoder} configuration using three different tracking tasks. The visual results of these tasks are provided in the supplementary video.

\paragraph{Unseen Motion} Firstly, we evaluate the tracking performance of \frameworkname{} on a test dataset that was not used during training. For this task, we extract a 2-hour dataset from HDM05 \cite{HDM05}, including activities such as walking, running, dancing, grabbing and various sports. We retarget these test motions onto our simulated character similarly to our training data processing. Then, we measure the Mean Per Body Position Error (MPBPE) between the simulated motion and the retargeted test motions as the performance metric. For our HDM05 dataset, the MPBPE of \frameworkname{} is 6.3\,cm. The simulated motion closely resembles the input, while artifacts from the motion retargeting, such as foot skating and penetrating the ground, are automatically fixed.

\paragraph{Kinematic Motion Generation}
Similar results can be observed when tracking the motions synthesized by a kinematic motion generator. To show this, we test two approaches:
\begin{itemize}
  \item \textbf{Latent Motion Matching}. We implement a simple motion matching strategy to generate continuous motion from an example motion clip. Specifically, we play the quantized latent codes of the motion clip in sequence. During playback, we monitor the $l_2$ distances between the current code and the codes of other frames in the motion clip. Once any of these distances falls below a threshold, we switch to a random qualified frame and continue playback from that point.
  \item \textbf{Latent Generative Model}. We utilize a pretrained motion latent diffusion model \cite{chen2023mld} as the motion generator. This model accepts text prompts and produces corresponding motion sequences, which are then retargeted onto our simulated character for tracking.
\end{itemize}

In the \emph{Latent Motion Matching} test, we evaluate our method on several dance motions. Using a single example dance clip, the character dances continuously, demonstrating diverse and non-repetitive dance patterns without noticeable artifacts. Unlike common motion matching implementations, such as DReCon \cite{bergaminDReConDatadrivenResponsive2019}, our approach does not need to blend animations in pose space, which typically requires careful design and tuning. The decoder of \frameworkname{} automatically handles random switches in playback and produces smooth motions, simplifying the process.

In the \emph{Latent Generative Model} test, we notice that even the state-of-the-art method \cite{chen2023mld} can generate motions containing unnatural shaking and floating movement. After retargeting and tracking with our method, these artifacts are effectively eliminated.

\paragraph{Video-based Pose Estimation}
Next, we show that the \emph{encoder-decoder} configuration of our \frameworkname{} can track results from monocular 3D human pose estimation. Due to the inherent ambiguity of the problem, these results often exhibit artifacts such as tilting and unbalanced posture, noisy and jittering movement, incorrect global position, and self-penetration, making them more challenging than the previously mentioned tasks. 

We employ an off-the-shelf method, HybrIK \cite{li2021hybrik}, as the human pose estimator. We apply this method to the test set of the Human3.6M dataset \cite{Human36MDataset} and use our \frameworkname{} to track the results. 
We compare this approach with the kinematic baseline HybrIK \shortcite{li2021hybrik} and several recent online physics-based human pose estimation methods \cite{PhysCap2020,SimPoE2021} on the same test set. It should be noted that these baseline methods are trained on the training set of Human3.6M, whereas our approach has never seen this dataset before. Additionally, previous physics-based methods often rely on residual force on the root of the character to maintain balance and global position. This mechanism is not physically plausible and can make the simulated character appear puppet-like. Our \frameworkname{} framework does not use any residual force and is physically accurate. 

\begin{table}[t]
  \centering  
  \caption{Evaluation on Human3.6M dataset. The baseline methods are HybrIK \cite{li2021hybrik}, PhysCap \cite{PhysCap2020}, and SimPoE \cite{SimPoE2021}.}
  \begin{tabular}{cccc}
      \hline
      \multirow[c]{2}{*}{Methods}
       & \multirow[c]{2}{*}{Physics}  & MPJPE $\downarrow$ & PA-MPJPE $\downarrow$ \\
       & & (mm) & (mm) \\
       \hline
      HybrIK & none & 54.4 & 34.5  \\
      PhysCap &  approx & 97.4 & 65.1\\
      SimPoE & approx & 56.7 & 41.6\\
      SimPoE {\small(w/o root force)} & full & 115.2 & 65.1\\
      \hline
      Ours  & full & 125.6 & 69.3 \\
      \hline
  \end{tabular}
  \label{tab:h36m:mpjpe}
\end{table}

\begin{table}[t]
  \centering    
  \caption{Motion quality evaluation on Human3.6M dataset}
  \begin{tabular}{cccc}
      \hline
      Methods & $e_{\text{smooth}} \downarrow$ &  $\sigma_{\text{smooth}} \downarrow$ & Accel $\downarrow$ \\
      \hline
      HybrIK & 5.9 & 3.1 & 10.9 \\
      PhysCap & 7.2 & 6.9 & -\\
      SimPoE &  -   &  -   &  6.7  \\
      SimPoE {\small(w/o root force)} & -&- & 23.5 \\
      \hline
      Ours & $\mathbf{3.4}$ & $\mathbf{2.9}$ & $\mathbf{5.1}$ \\
      \hline
  \end{tabular}
  \label{tab:h36m:physical}
\end{table}

Table~\ref{tab:h36m:mpjpe} provides a qualitative comparison with these baseline approaches. We evaluate the Mean Per Joint Position Error (MPJPE) and the Procrustes-Aligned MPJPE (PA-MPJPE) on the simulated motion, which are widely used metrics to evaluate accuracy in human pose estimation. We scale our simulated motion to match the height of the test subjects, S9 and S11 of Human3.6M, for a fair comparison. In Table~\ref{tab:h36m:mpjpe}, our method achieves results comparable with the baseline physics-based methods, whether in the approximate or full-physics configurations, on the PA-MPJPE metric without any training or fine-tuning on the Human3.6M dataset. As a simple test, we did not implement global position and orientation control in this tracking task. As a result, the MPJPE metric without Procrustes alignment is higher but remains comparable to SimPoE \cite{SimPoE2021} under the full-physics setting. 

Table~\ref{tab:h36m:physical} compares the motion quality of these methods. We evaluate the smoothness indicator $e_{\text{smooth}}$ and its standard deviation $\sigma_{\text{smooth}}$ as described in \cite{PhysCap2020}. Additionally, we report the difference in joint positional acceleration (Accel) between the predicted and real data, which serves as a measure of the motion's jitteriness. Qualitatively, although the kinematic baseline method HybrIK~\shortcite{li2021hybrik} exhibits the highest accuracy, its estimation for pose and global position is often noisy and physically implausible. Our tracking control serves as a post-processing operation, resulting in clean and high-quality motions. 

\fig\ref{Fig:rvq} shows several snapshots of our tracking results. The simulated character closely mirrors the human motion in the video. Notably, in complex scenarios such as arm crossing, the simulated motion remains physically correct and avoids self-penetration. We encourage readers to view the supplementary video for a better evaluation.

\begin{figure}[t]
    \centering
    \includegraphics[width=\linewidth]{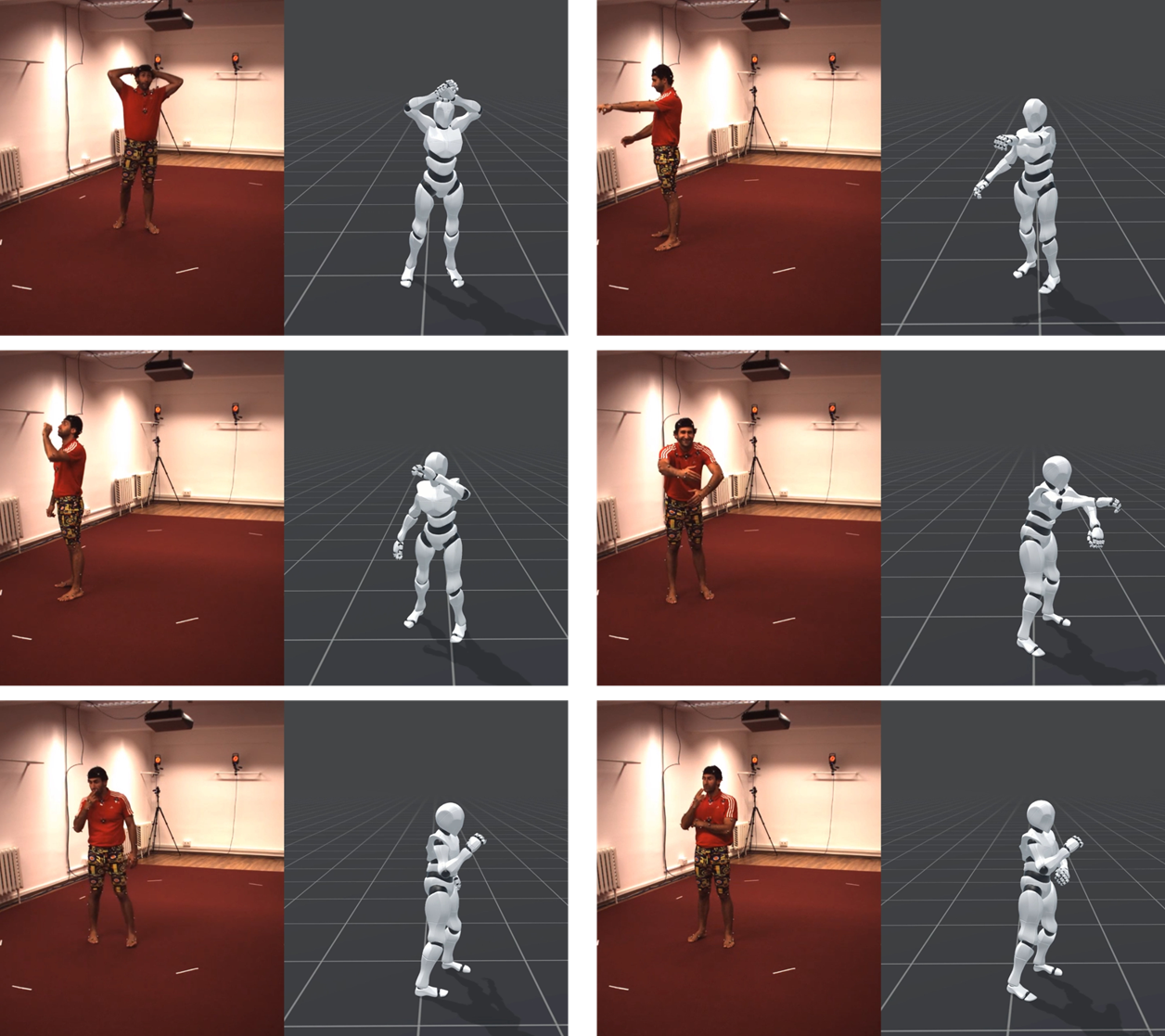}
    \caption{Physics-based human pose estiamtion by tracking the results of HybrIK \cite{li2021hybrik}.}
    \label{Fig:pose_estimation}
\end{figure}

\begin{figure*}[t]   
  \includegraphics[width=\linewidth]{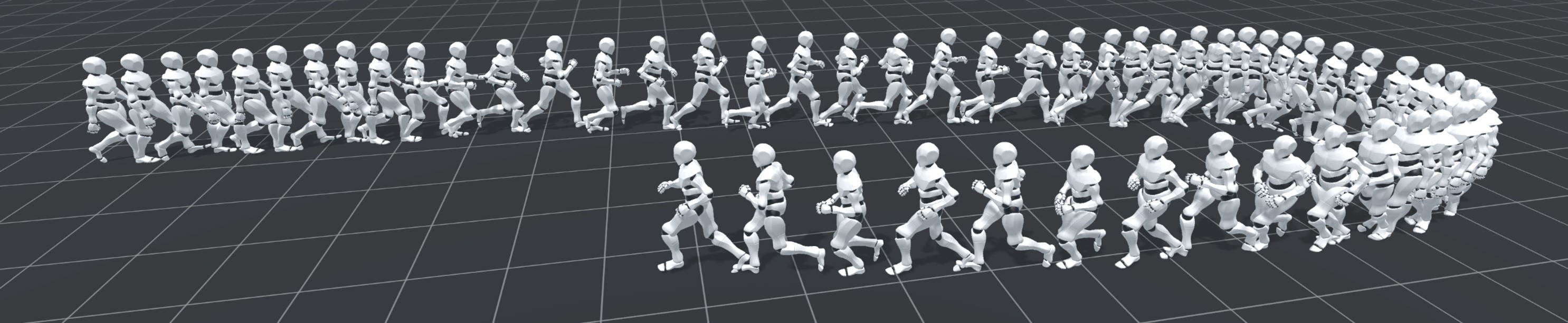}
  \caption{Walk and run under user control.}
  \Description{}
  \label{fig:user_control}
\end{figure*}

\subsection{Interactive Control}

We train our interactive control by distilling walking and running motions. A training dataset, comprising $8$ minutes of motion, is extracted from the LaFAN dataset \cite{harvey2020robustLafanYHY} for this task. We use the layer-wise mixture of experts (MoE) network from \citet{xieSoccer2022}. This network has $4$ experts, each being a $4$-layer MLP containing $256$ units per layer. The gating network is a two-layer MLP with 64 units. We employ the RAdam optimizer \shortcite{liu2019radam} with the learning rate set to $2\times 10^{-4}$. At runtime, users can interactively control the character using either a keyboard or a joystick. Their inputs are mapped to a desired character trajectory, which is then combined with previous network outputs as per \eqn\ref{equ:blend}. This desired trajectory is computed by blending the current velocity with the input velocity, leveraging the inertialization technology \cite{bollo2016inertialization}.

\begin{figure}[t]
  \centering
  \includegraphics[width=\linewidth]{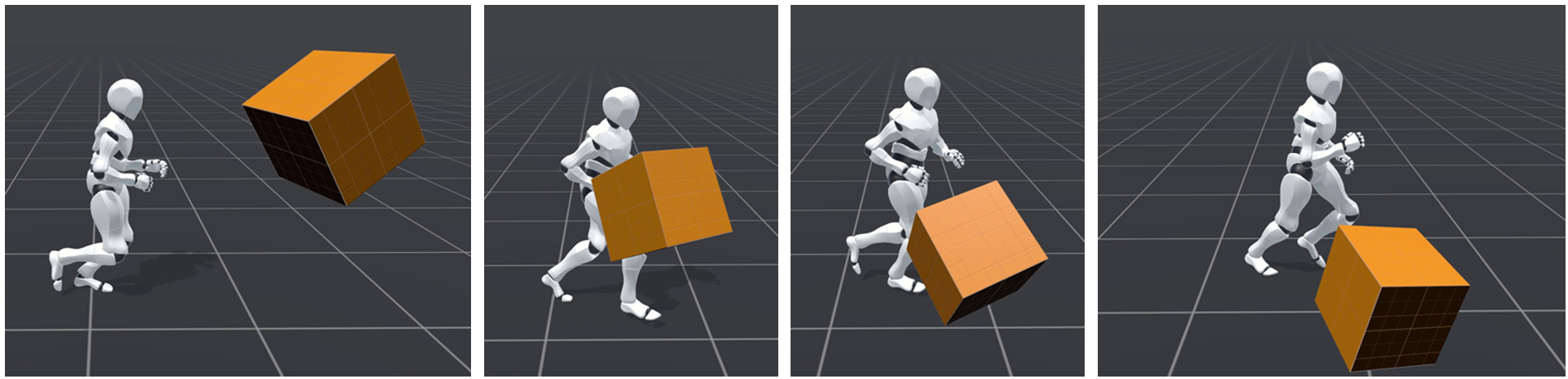}
  \caption{Recover after being hit by a box during interactive control.}
  \label{Fig:obstacle}
\end{figure}

\begin{figure}[t]
    \centering
    \includegraphics[width=\linewidth]{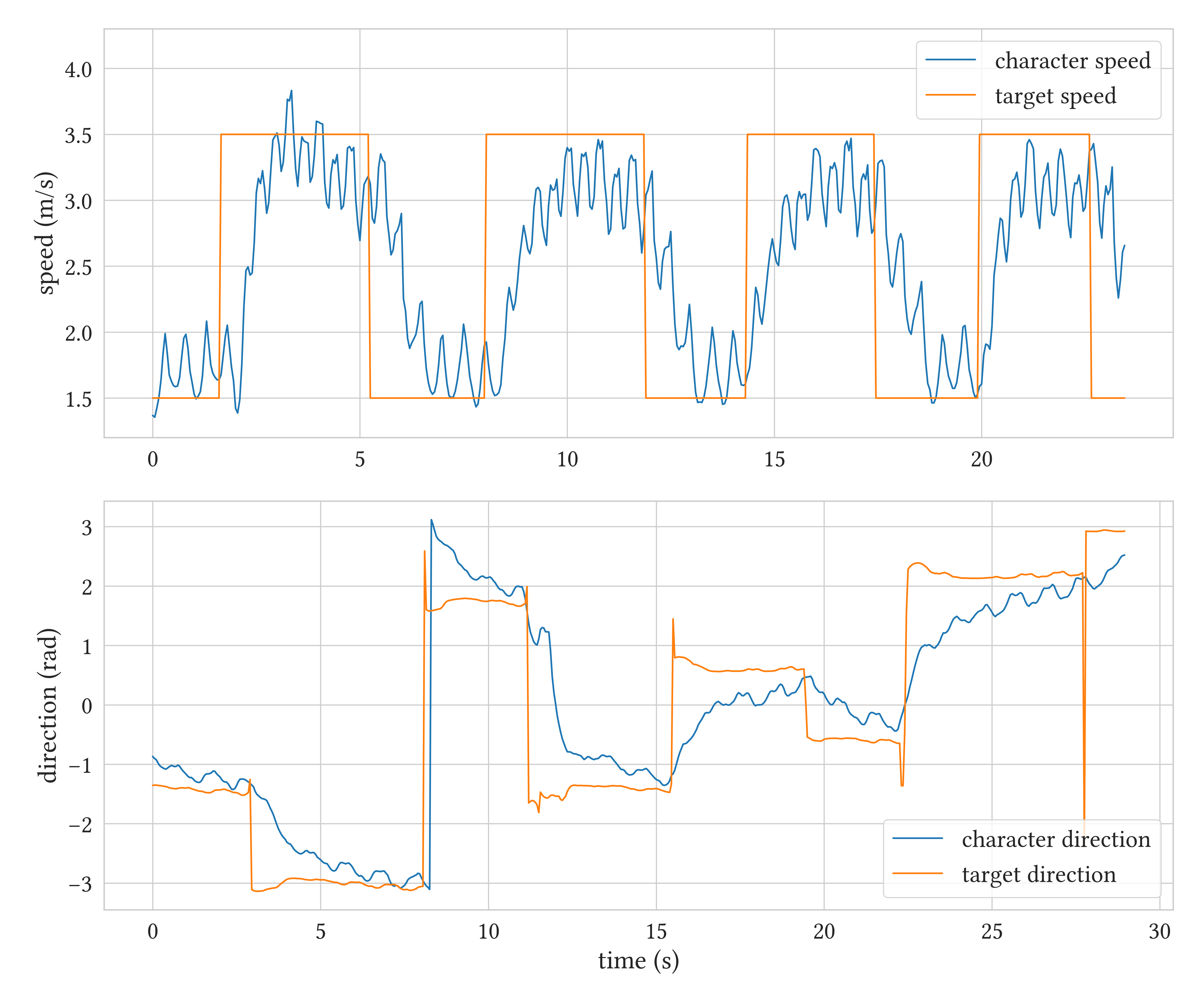}
    \caption{Control signals and responses. The control signals are input using a joystick, allowing the user to set the target speed and direction of movement. The upper portion of the figure shows the change in the character's speed in response to user input. The lower portion illustrates the change in direction of movement following user input.} 
    \label{Fig:respond}
\end{figure}

\fig\ref{fig:user_control} shows snapshots of character motions under joystick control, while \fig\ref{Fig:respond} provides a quantitative evaluation of our character's response to user input. Under user control, the character adeptly manages walking, running, and smooth transitions between these gaits. It can also smoothly handle varying directional inputs.
As a physics-based motion generation method, our approach is inherently capable of responding to external physical interferences while maintaining a high level of robustness. For example, as shown in \fig\ref{Fig:obstacle}, we can throw a box to the character from a random direction and speed when it is running. The character automatically adjusts its posture after experiencing the collision to maintain balance. If knocked down, the character can automatically rise from the ground and resume running.


\begin{figure}[t]
  \centering
  \includegraphics[width=\linewidth]{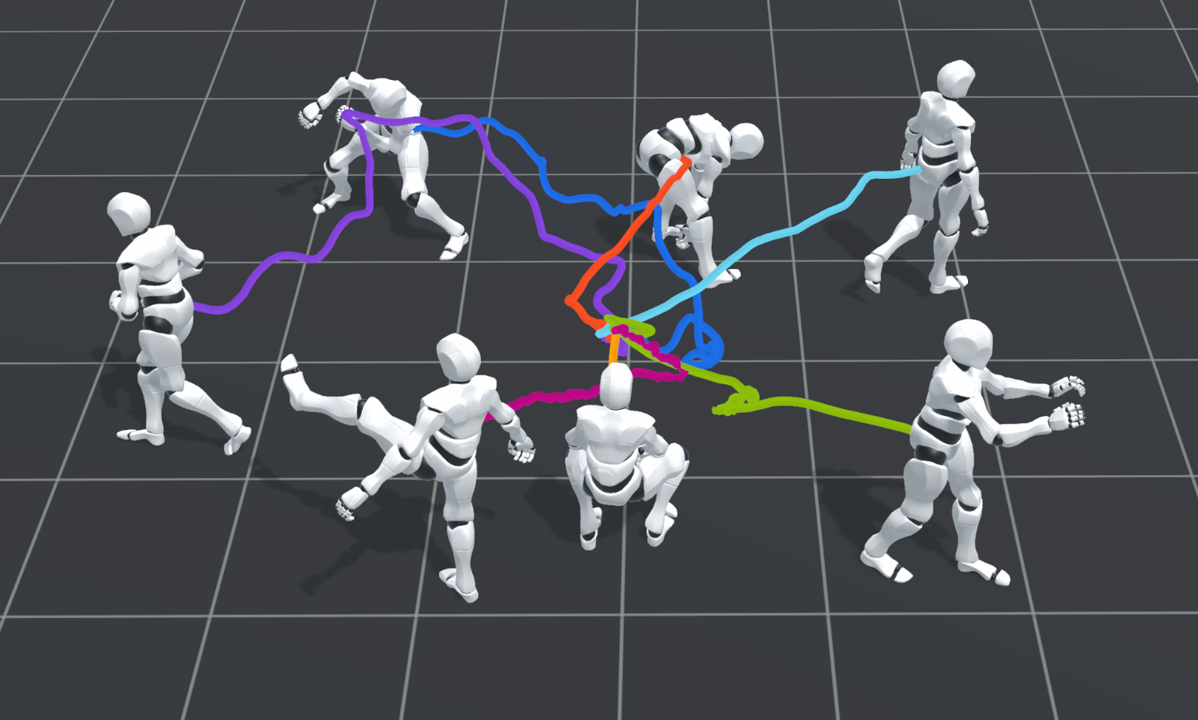}
  \caption{Unconditional motion generation with \controlGPT{}. All characters are initialized with the same position and state. With random sampling, the characters perform smooth motions with various types and trajectories.}
  \label{Fig:unconditional_generation}
\end{figure}
\subsection{Generative Motion Prior with \controlGPT{}}

\begin{figure*}[t]   
 \begin{subfigure}{\linewidth}
    \includegraphics[width=\linewidth]{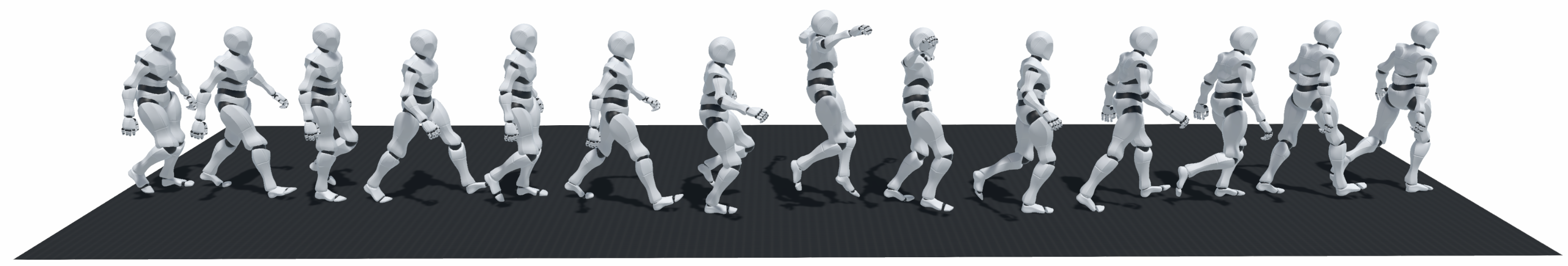}
    \caption{"a man walks forward and jumps and walks forward"}
  \end{subfigure}

  \begin{subfigure}{\linewidth}
    \includegraphics[width=\linewidth]{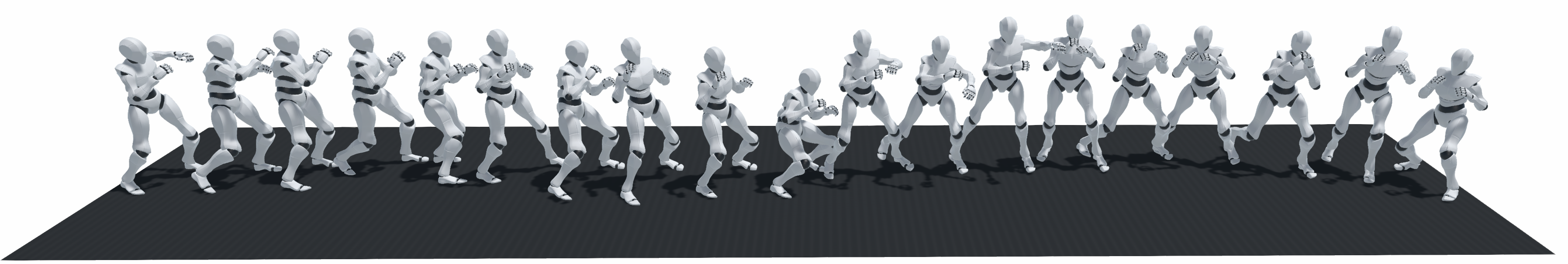}
    \caption{"a person is boxing and jumping"}
  \end{subfigure}
  
  \begin{subfigure}{0.41\linewidth}
    \includegraphics[width=\linewidth]{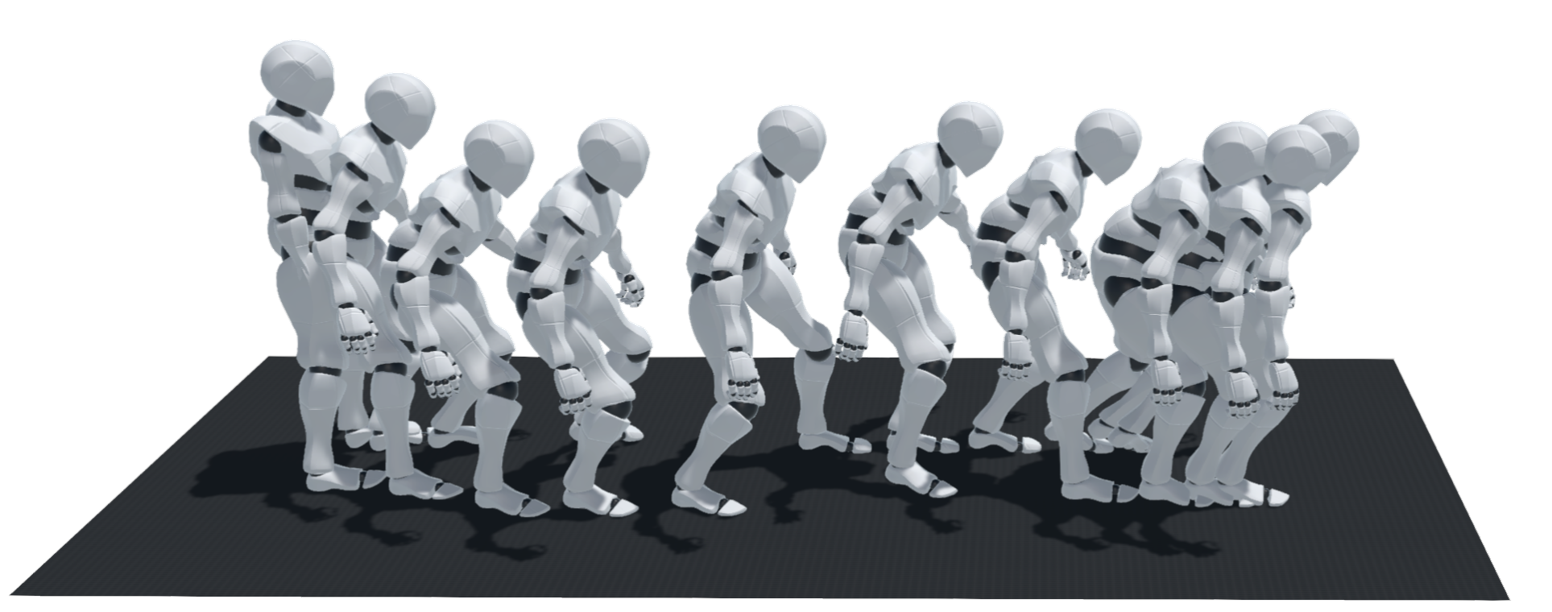}
    \caption{"a person slightly crouches down and walks forward, then he stands still"}
  \end{subfigure}
  \begin{subfigure}{0.58\linewidth}
    \includegraphics[width=\linewidth]{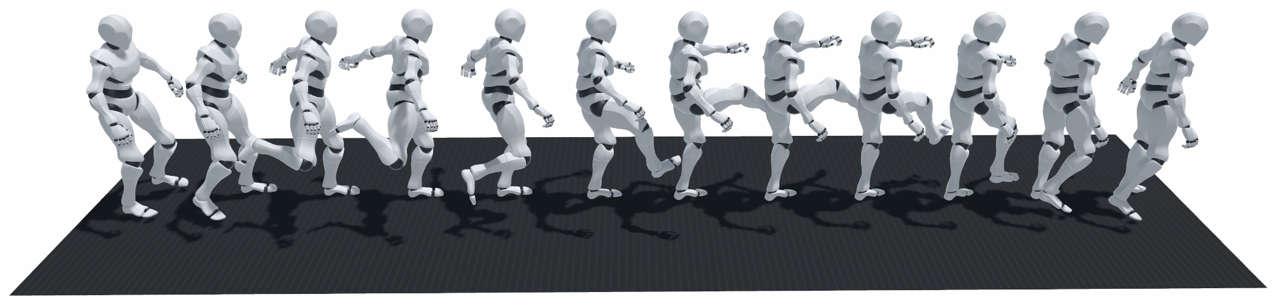}
    \caption{"a man is kicking with right leg"}
  \end{subfigure} 
  
  \Description{}
  \caption{Physics-based text-to-motion generation using T2M-\controlGPT{}.}
  \label{fig:simulated_skills}
\end{figure*}

We train our physics-based \controlGPT{} on the HumanML3D dataset \cite{hml3d}, which is a 3D human motion dataset with textual descriptions. After filtering and mirror augmentation, we obtain approximately 20k text-motion pairs. To prepare the training data, we encode the motions into latent embeddings and indices using our \frameworkname{} encoder. Text features are extracted using the pretrained T5 encoder \cite{raffel2020exploringt5}. As for the parameter settings, our temporal transformer has $N_T=12$ attention units, while the depth transformer has $N_D=5$ attention units.  The entire \controlGPT{} model accepts and generates motion code sequences up to $K=50$ frames, corresponding to a motion of $10$ seconds. We employ the RAdam optimizer \cite{liu2019radam} with a learning rate set to $2\times{}10^{-4}$, and the training process lasts for 150k iterations.

\paragraph{Unconditional Generation}
We first demonstrate unconditional motion generation using our \controlGPT{}. The text feature and the corresponding cross-attention layers are removed from the structure shown in \fig\ref{Fig:transformer}. We utilize a random Gaussian noise vector as the starting code for the temporal transformer and run \controlGPT{} in an autoregressive manner. Subsequent motion codes are sampled from the conditional distribution defined in Equation \eqref{equ:control_gpt}. \fig\ref{Fig:unconditional_generation} presents several sampled motion sequences from this unconditional generation process. As shown, the generated motions cover a diverse range, from basic locomotion activities like walking and running to specialized tasks such as dancing, sitting on a chair, and balancing on a single leg.

Although our \controlGPT{} model accepts motion sequences of limited length, we can easily extend it using sliding windows. We take the last $k=5$ codes generated by the previous step and use them as the first $k$ codes for the next step. By doing so, we can generate arbitrarily long sequences. Results show that the generated motions maintain smoothness at the boundaries of each segment.

\paragraph{Text2Motion Generation}
We train our T2M\hyp{}\controlGPT{} model with the text features enabled as shown in \fig\ref{Fig:transformer} to generate motions from text descriptions. \fig\ref{fig:simulated_skills} displays several results from random text inputs. Our method understands and responds to the input text prompt, generating physically accurate motions. The character performs action sequences described in the input sentences smoothly, even when such action combinations do not exist in the training dataset.

As for a quantitative evaluation, we compute the FID and R-precision metrics for the data generated by our method on the test set of HumanML3D \cite{hml3d}. These metrics are often used for assessing text2motion algorithms. Table~\ref{tab:text2motion} provides a comparison with several baseline approaches \cite{hier,tevet2023mdm,t2mt,t2mgpt,PADL}. It is worth noting that our method is the first physics-based approach to address a general text-to-motion task. Our results have comparable performance in terms of semantic match (R-precision) and motion similarity (FID), and do not exhibit artifacts like floating, penetration, or jittering, which commonly exist in the baseline kinematic approaches. For a fair comparison, we utilize the same pretrained motion feature extractor \cite{hml3d} as previous works to compute the metrics. This extractor is trained on the SMPL skeletal framework \cite{SMPL2015Model}, which requires motion retargeting between our simulated character and SMPL. However, our basic copy-joint-rotation retargeting strategy was found to significantly impact the metric values, explaining the relatively low R-precision of our method. To illustrate the influence of this motion retargeting, we also present the R-precision of the ground-truth motion after it has been retargeted back and forth between our simulated character and SMPL. This clearly highlights the decrease in R-precision following the retargeting.

\begin{table}
\centering
\caption{Comparison with the state-of-the-art text-to-motion generation methods. We compute the metrics following \cite{hml3d} on the test set of HumanML3D. Metrics for Hier \cite{hier}, MDM \cite{tevet2023mdm}, TM2T \cite{t2mt}, and T2MGPT \cite{t2mgpt} has been reported in their papers. 
To demonstrate the effect of the retargeting process, we retarget the ground truth motion to our character and then retarget it back, the results are shown in the row of GT (Retargeted). 
}
\label{tab:text2motion}

\begin{tabular}{c c c c c c}
\hline
\multirow{2}{*}{Model} & \multicolumn{3}{c}{R-precision \  $\uparrow$} & \multirow{2}{*}{FID \ $\downarrow$} & \multirow{2}{*}{ \makecell[c]{ Physically \\ Accurate} }\\
\cline{2-4}
 & Top 1 & Top 2 & Top 3 & & \\
\hline

Hier  & 0.301& 0.425& 0.552& 6.532 & \usym{2717} \\
MDM  &0.320 & 0.498 &0.611 & 0.544 & \usym{2717} \\
TM2T  & 0.424& 0.618& 0.729& 1.501 &\usym{2717} \\
T2MGPT  & 0.491& 0.680& 0.775& 0.116& \usym{2717} \\
\hline
Ours &0.367 &0.561 &0.678 & 0.254 &\usym{2713} \\
\hline
GT (Original) & 0.511 & 0.703 & 0.797 & - & -  \\
GT (Retargeted) & 0.440 & 0.639 & 0.753 & - & - \\
\hline
\end{tabular}

\end{table}

\subsection{Integration with Large Language Model}
\label{sec:chatgpt}

In this section, we test the integration of our \frameworkname{} with commercial large language models. The in-context learning framework requires examples to be given in text, which means the token limit of an LLM becomes a critical factor. This limit constrains the volume of examples and can potentially affect learning performance. While ChatGPT-4 \cite{chatgpt} is one of the most popular LLMs, it only accommodates a token limit of either 8k or 32k, depending on the version. Claude-2 \cite{claude}, another well-known commercial LLM, supports a much larger context size of 100k. This extended capacity is particularly beneficial because it allows for more extensive demonstrations, which can enhance the model's understanding and adaptability. Consequently, we chose Claude-2 in this task.

We provide Claude-2 with 1600 examples of paired text descriptions and index-based representation of motions, which are randomly selected from the HumanML3D dataset \cite{hml3d}. On average, each example takes 167 characters. \fig\ref{fig:prompt} illustrates how we use a text prompt to guide Claude-2 in learning the motion representation. Note that Claude-2's interface allows for the uploading of files containing text prompts. Therefore, we place the motion examples in a file and uploaded it alongside the other text prompts.

\begin{figure}[t]
    \centering
    \small
    \begin{tabular}{p{0.9\columnwidth}}
    \toprule[1pt]  
    \textbf{Prompt:} Learn Motions from Text \\ 
    \midrule[1pt]

    \begin{list}{}{
      \setlength{\itemsep}{0.5em}
      \setlength{\leftmargin}{0em}
      \setlength{\rightmargin}{0em}
      } 

    \item Here's a representation of motion. We will provide you with a description of the motion, followed by a discrete representation in brackets. 

    \item {[File]}

    \item Please learn from this, and then, if I provide a motion description,respond with the index sequence. 
    
    \item For instance, if I say, `A man stretches his hands to the left, picks something up, tilts to the right, and then places it there', you should respond with [503,503,416,.., 386]. 
    \end{list}    \\
    \bottomrule[1pt]
    \end{tabular}    
    \caption{Prompt used for in-context learning with an LLM.}
    \Description{}
    \label{fig:prompt}
\end{figure}

\begin{figure}[t]
    \centering
    \small
    \begin{tabular}{p{0.9\columnwidth}}
    \toprule[1pt]
    ...\\
    \textbf{User:} Question: `a person walks backwards and kicks'\\ 
    \textbf{LLM:} 
    \begin{enumerate}
        \item "walks backwards": ... 
        \item "kicks": Represented by "110 273 67 178 299 56 419 504 386 110 437"
    \end{enumerate} \\
    \textbf{User:} Could you explain how do you extract the kick sequence? \\ 
    \textbf{LLM:}
    I chose this subsequence because:
    \begin{enumerate}
        \item 110 at the start represents preparation for the kick.
        \item 299 and 56 capture the key kick out motion itself.
        \item 419 and 504 represent the follow through.
        \item 110 and 437 at the end reset the position after the kick.
    \end{enumerate} \\
    \textbf{User:} Why do you think 299 and 56 capture the key kick out motion? \\ 
    \textbf{LLM:} 
    \begin{enumerate}

    \item In the examples, 299 is used in sequences that involve motions like kicking or extending the leg outwards in a forceful manner.
    
    For example:
    \begin{enumerate}
        \item "a person kicks something with their right foot" uses 299.
        \item "a person is performing a right leading front kick" uses 299."
    \end{enumerate} 

    This suggested to me that 299 captures a kicking or leg extension motion.
    
    \item 56 is also used alongside 299 in representing kicks or foot strikes, like: "a person turns a half circle counter clockwise and moves arms in a sweep motion from left to left while crouching down a little bit" uses 299 and 56 together. This coupling suggested 56 also captures part of the kick motion.

    \item The order 299 followed by 56 flows well - 299 feels like the wind up or preparation for extending the leg, while 56 feels like the follow through.

    \item So in summary, based on how 299 and 56 were used in the examples, they seemed to represent:
        \begin{itemize}[label={}]
        \item 299: Extending or kicking out the leg
        
        \item 56: The follow through after the kick
        \end{itemize}
    \end{enumerate}
    \\
    \bottomrule
    
    \end{tabular}
    
    \caption{Instruct an LLMs to explain the reason for choosing specific indices.}
    \Description{}
    \label{fig:prompt_explained}
\end{figure}

\begin{figure*}[t]
  \includegraphics[width=\linewidth]{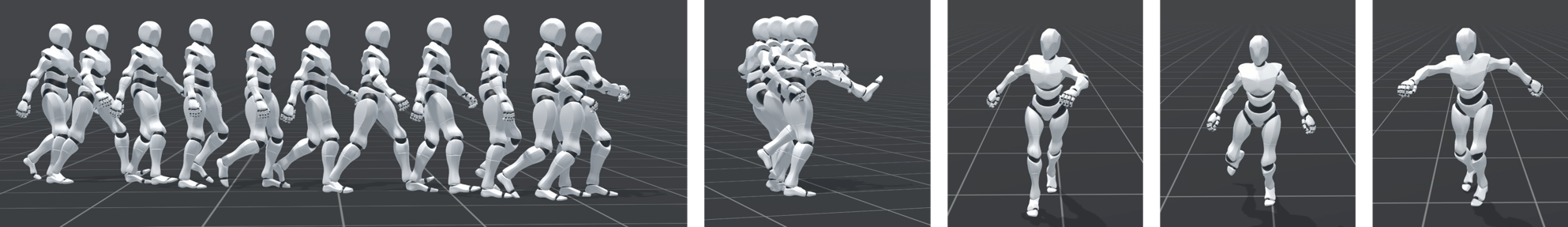}
  \caption{Generating a long motion using an LLM. The input description is: \textPrompt{a person walks forward for a long time and kicks, then he begins to dance}. The LLM extracts skills from the provided examples and combines them to match the given description.}
  \label{fig:claude_gen}
\end{figure*}

\paragraph{Text2Motion with LLM}
A direct application of this integration is to use LLM to generate new motions according to descriptions in natural language. \fig\ref{fig:claude_gen} shows an example, where the motion is generated by the LLM in response to our text prompt: \textPrompt{a person walks forward for a long time and kicks, then he begins to dance}. The LLM automatically extracts the index sequences that represent the actions \emph{walk forward}, \emph{kick}, and \emph{dance}. To satisfy the condition \textPrompt{for a long time}, it repeats the indices corresponding to \emph{walk forward} several times. Additionally, to capture the \emph{kick} action, the LLM identifies repetitive elements from another example phrase, \textPrompt{a person kicks their right foot twice}, enabling it to isolate the single \emph{kick} movement and integrate it into the generated sequence. Interestingly, the LLM can, when prompted, provide insights into how it interprets a given example. An illustration of this capability is shown in \fig\ref{fig:prompt_explained}. More examples can be found in our supplementary video.

The LLM can occasionally exhibit suboptimal actions when misinterpreting the examples or the new query. For example, it may break down the examples into excessively small snippets and treat a single index as an action. It can also be confused by an imperfect motion description. Moreover, it may cheat by finding an example whose description is semantically similar to the query and then directly outputting that example motion. If this occurs, we provide further prompts to guide its correction. For instance, we input: \textPrompt{action clips like "walk forward" can't be represented by just a single number, and the text annotations aren't flawless, the action sequence might include some preparatory movements, so don't use the entire example sequence without analysis} as the additional prompt, the LLM can refine its interpretation and produce a more accurate and relevant motion outputs.

\begin{figure}[t]
  \centering
  \includegraphics[width=0.9\linewidth]{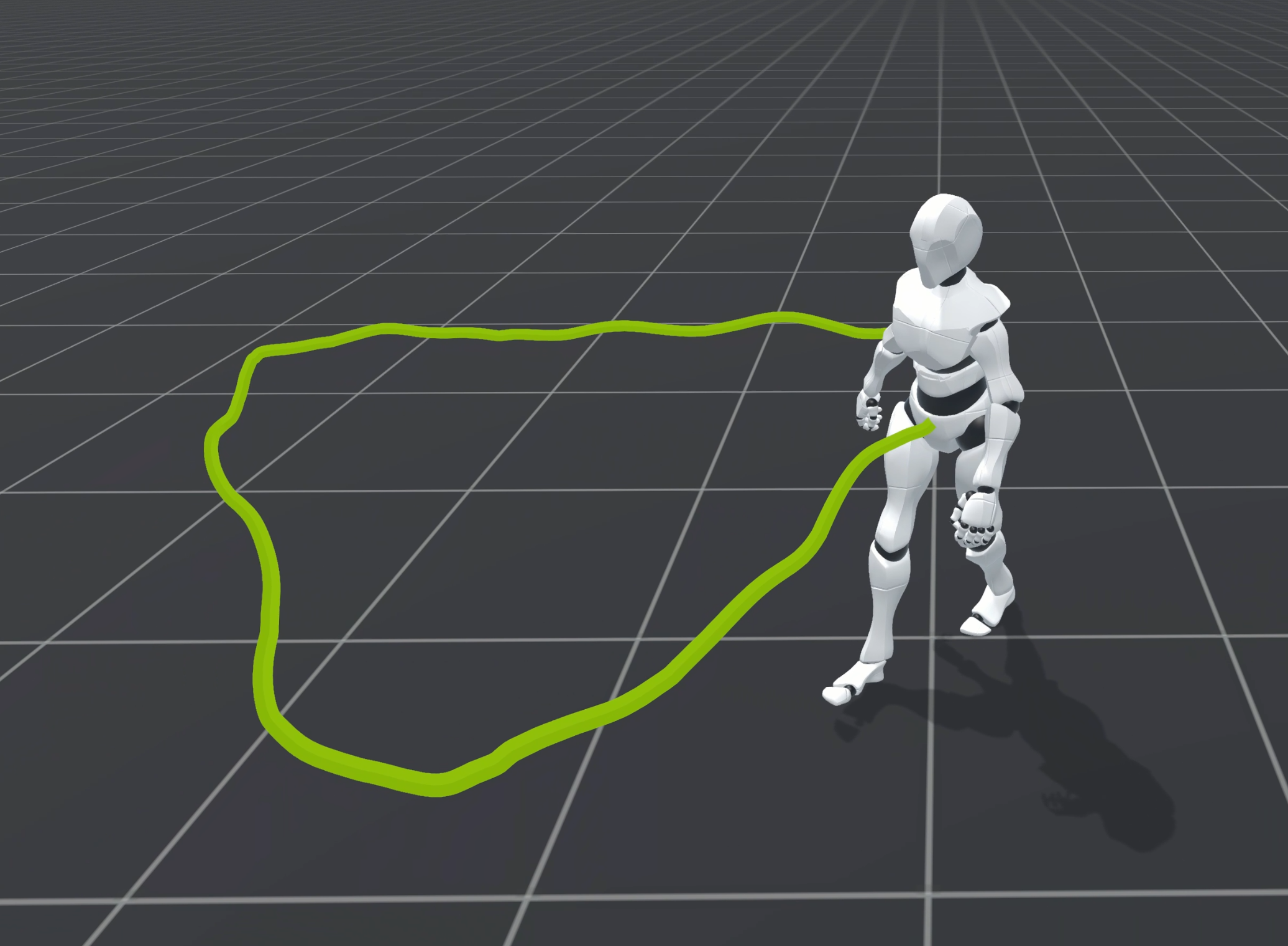}
  \caption{LLM's response to the abstract task: \textPrompt{tell me how to control the character walking with a square trajectory}. The character walks and makes four turns, though the trajectory resembles a triangle due to the lack of angle information.}
  \label{Fig:square}
\end{figure}
\paragraph{Abstract Tasks with LLM}
The reasoning capability of LLM enables it to handle more complex and abstract tasks using learned actions, rather than just translating motion descriptions. For example, when we instruct the LLM with \textPrompt{tell me how to control the character walking with a square trajectory}, the LLM recognizes that the character must repeat a forward-moving action and a turning action four times, and then generates an index sequence combining these actions. In contrast, our T2M-\controlGPT{} cannot correctly interpret such an abstract description, as no such example exists in its training dataset. \fig\ref{Fig:square} showcases the generated results. The character performs four \emph{walk forward} and \emph{turn} actions as expected. However, its trajectory more closely resembles a triangle than a square, as the example text annotations do not specify the turning angles.

\begin{figure}[t]
  \centering
  \includegraphics[width=\linewidth]{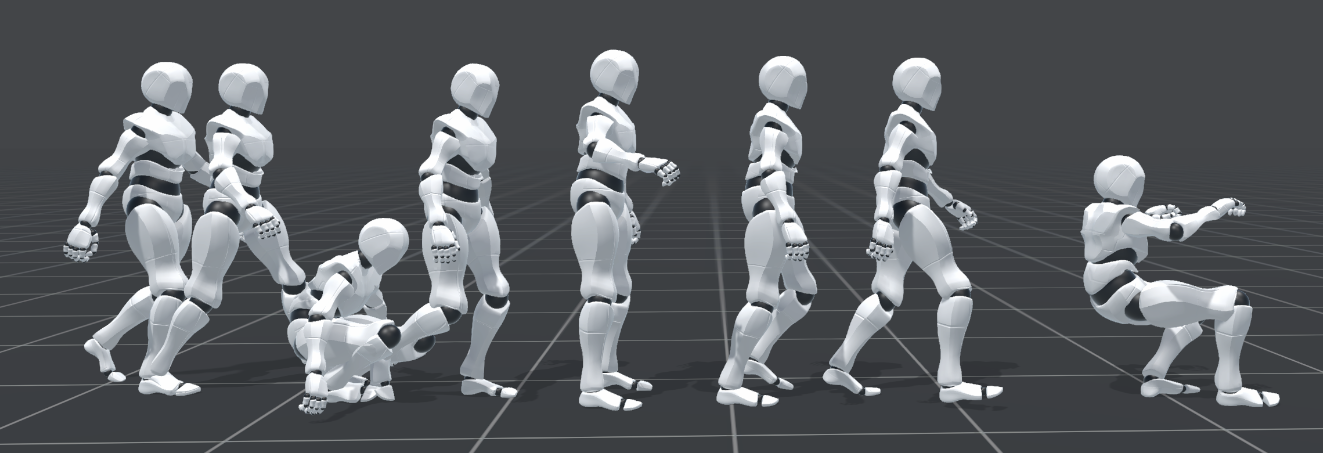}
  \caption{LLM's response to a task with a scene description and an abstract goal. Please see the text for details.} 
  \label{fig:key}
\end{figure}

Another example, as shown in \fig\ref{fig:key}, depicts a more complex task. We prompt the LLM with a description of a scene and then ask it to complete a task within that scene. The query is:

\begin{minipage}[c]{0.9\linewidth}
\vspace{0.5em}
  \textPrompt{Question: Suppose there is a door in front of you and a key on the ground. you want to open the door and sit on the chair, how do you perform? Please describe how you get index sequences for each action.}
\vspace{0.5em}
\end{minipage} 

\noindent
The LLM then instructs the simulated character to walk to the key, pick it up, stand up, unlock and open the door, and finally sit down. Interestingly, there is not an \emph{open a door} action in the provided examples, but the LLM extrapolates using the movement from the \emph{cutting beef with a knife} example to accomplish the task. The complete response of the LLM is provided in the appendix.

\subsection{Ablation Studies}
\label{sec:ablation}
In this section, we conduct experiments to demonstrate the effectiveness of two key designs in our \frameworkname{} framework: vector quantization and the residual structure. We evaluate these components using the \emph{encoder-decoder} configuration, employing the 2-hour test dataset extracted from the HDM05 \cite{HDM05}, as introduced in Section~\ref{sec:tracking_results}, for this evaluation.

\begin{figure}[t]
  \centering
  \begin{subfigure}{\linewidth}
    \includegraphics[width=\linewidth]{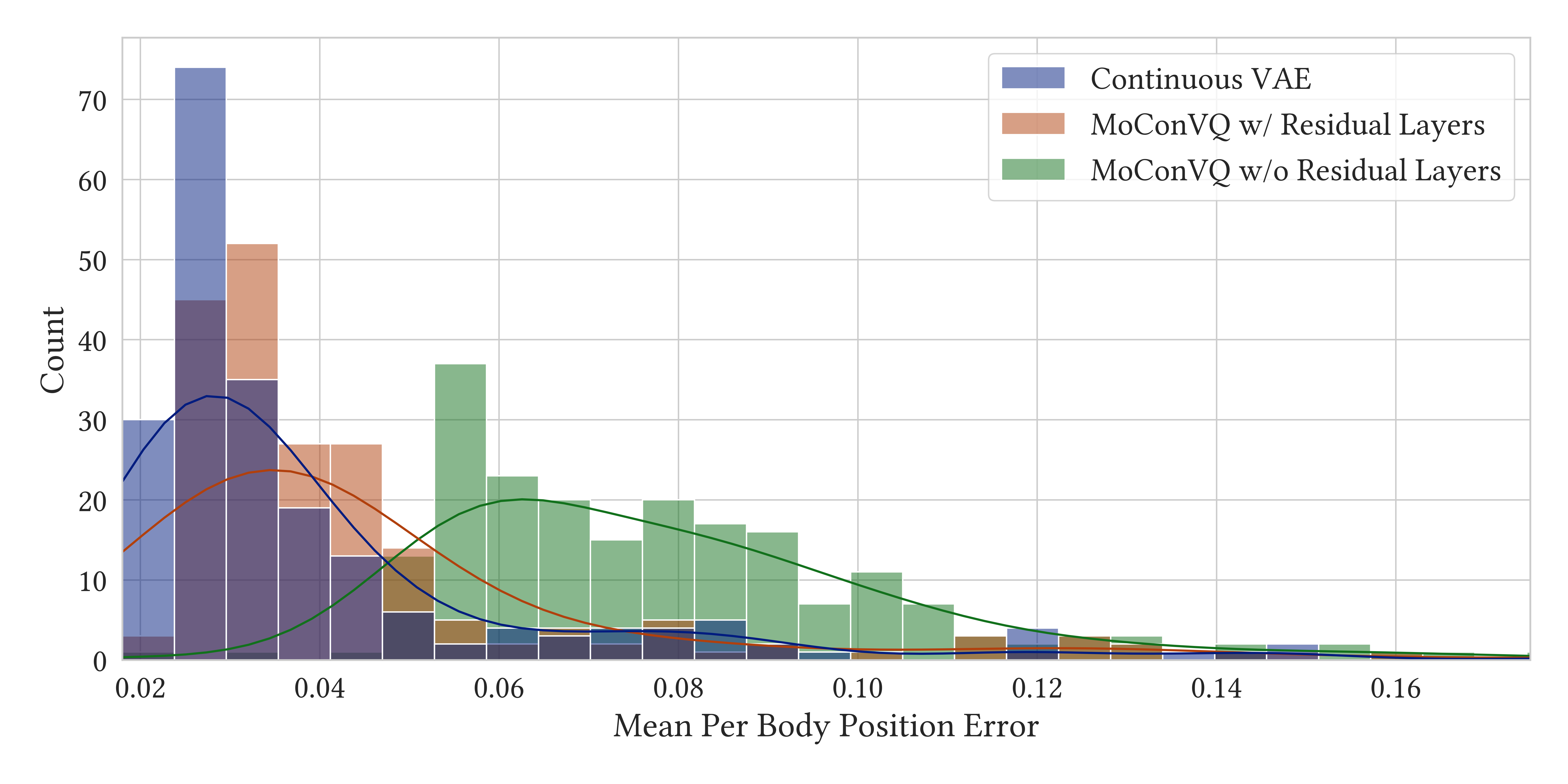}
    \Description{}
    \caption{Clean motion data}
  \end{subfigure}
  \begin{subfigure}{\linewidth}
    \includegraphics[width=\linewidth]{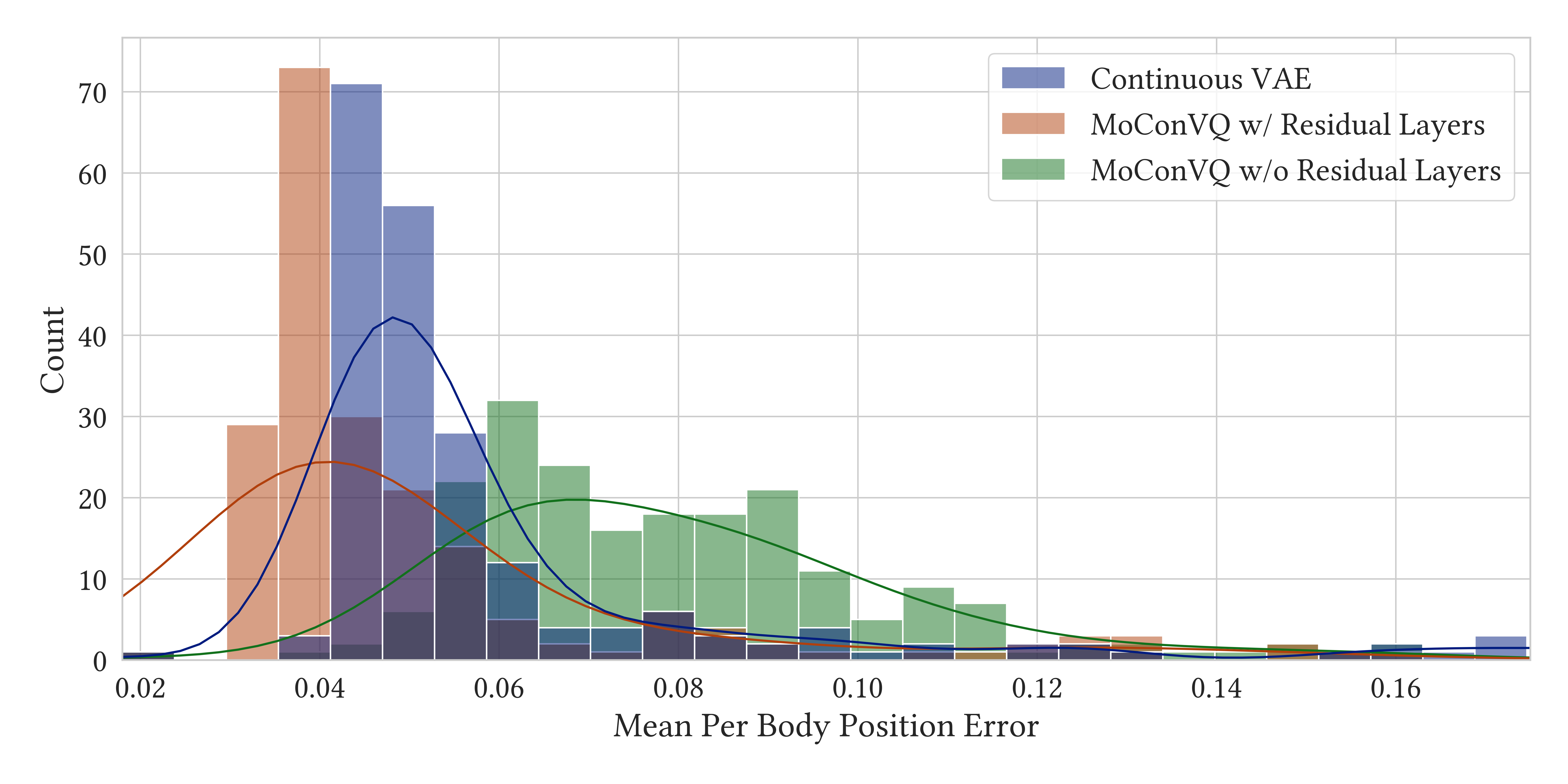}
    \Description{}
    \caption{Motion data corrupted by adding Gaussian noise $\epsilon \sim \mathcal{N}(0,0.1)$}  
  \label{fig:seed_rwd}
  \end{subfigure}
  \caption{Tracking error of different motion representations on the same test set from HDM05 \cite{HDM05}. We test three models: i) The continuous VAE model with KL regularization, ii) Our \frameworkname{} model with $8$ VQ layers, and iii) Our \frameworkname{} model with only the VQ layer. The curves represent kernel density estimations.}
  \Description{}
  \label{Fig:Tracking_Error}
\end{figure}

\paragraph{Vector Quantization}
First, we evaluate the effect of the vector quantization of \frameworkname{}. As a baseline, we employ a continuous VAE structure that has the same architecture as our \frameworkname{} but omits the quantization in \eqn\eqref{equ:encode_decode}. This continuous VAE is trained in the same manner as \frameworkname{}, except that the commitment loss and the VQ objective in the training loss \eqn\eqref{equ:vq_vae_loss} are replaced by a KL divergence bottleneck.

We compare the tracking performance of both the continuous VAE and \frameworkname{} on the test dataset. The results are depicted in \fig\ref{Fig:Tracking_Error}. It can be seen that the continuous VAE exhibits superior accuracy when tracking clean test motions. This is expected, as it is less constrained by the regularization imposed by the vector quantization operation. However, when the test motions are interrupted by introducing moderate Gaussian noise $\epsilon\sim\mathcal{N}(0, 0.1)$ to the joint rotations, the performance of the continuous VAE drops dramatically. In contrast, the performance of \frameworkname{} remains largely unchanged. This suggests that vector quantization is efficient at countering perturbations, leading to robust motion tracking across various sources

\begin{figure}[t]
  \centering
  \includegraphics[width=\linewidth]{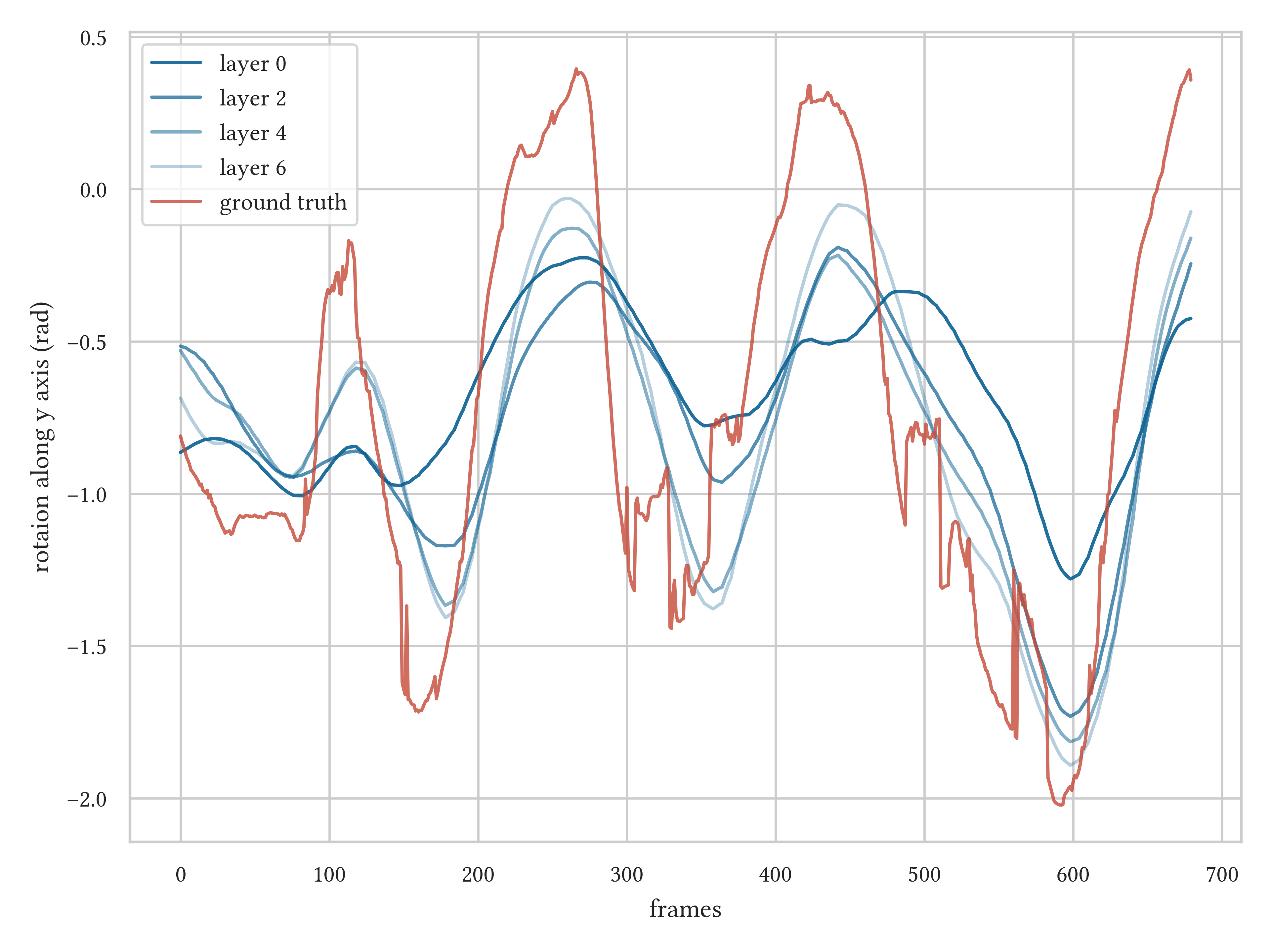}
  \caption{Motion tracking with an increasing number of RVQ layers. The ground truth is a noisy dance motion from the test dataset. The curves visualize the rotation of the left elbow.} 
  \Description{}
  \label{Fig:rvq}
\end{figure}

\paragraph{Residual VQ}
Next, we evaluate the effect of the residual structure of \frameworkname{} in the same motion tracking test. As revealed in \fig\ref{Fig:Tracking_Error}, the adoption of the residual layers significantly enhances motion-tracking precision while retaining the robustness of vector quantization.
Additionally, we visualize the performance of different numbers of residual layers in \fig\ref{Fig:rvq}. For this test, we use a noisy dance motion that was not part of the training dataset. It can be seen that the VQ layer captures the overall characteristics of the motion trajectory, whereas the subsequent residual layers refine the finer details.

\section{Conclusion}

In this work, we have presented \frameworkname{}, a unified and scalable framework for physics-based character animation. We show that the discrete motion representation learned via VQ-VAE can effectively accommodate a wide variety of motions. Additionally, a residual structure compensates for missing motion details due to quantization, ensuring high-quality motion generation. Through our model-based RL approach, the model can process large-scale motion data spanning more than twenty hours and effectively encode the actions and transitions within that data. We demonstrate that this motion representation, paired with our physics-based encoder and decoder, serves as a unified interface for a variety of downstream applications, including interactive motion control, physics-based monocular 3D human pose estimation, text-to-motion generation, and, more interestingly, integration with large language models.

Our framework is not without limitations. First, while our model-based RL training algorithm can handle a diverse range of motions, we find it challenging to learn specific actions, such as backflips and certain kung fu movements, which are sparse in the dataset and isolated from other motions. A potential solution might be the adoption of a progressive reinforcement learning method, like the progressive network \cite{PerpetualHC,Rusu2022PNN}. Second, our convolutional network-based decoder requires several future motion codes to compute the current action, which could introduce latency in real-time applications. A potential workaround might involve using a causal network structure, such as causal convolutions \cite{Oord2016WaveNet}, for both the encoder and decoder. Third, our motion representation does not differentiate  between the body parts of the character, making it challenging to control individual limbs directly. Exploring combinatorial motion representations, as demonstrated by recent works \cite{Ao2023GestureDiffuCLIP,PMP,compositeGAN,Lee2022bChimeras}, could be interesting. Lastly, our current framework is designed for single-person actions. Expanding it for interactions with the environment and multi-agent applications would be a valuable direction for future research.

We utilize several applications to showcase the potential of the unified motion control framework provided by our method. While the design of these applications is not optimized meticulously, they already achieve results comparable to state-of-the-art approaches. For future work, we wish to explore more tasks that can be integrated with our framework, such as generating physics-based dance and gesture using multi-modal information. Integrating environmental feedback from the character's actions into the large language model, enabling it to solve abstract tasks through closed-loop control, is another promising avenue. We hope our work paves the way for increased exploration of large-scale data and models in the field of physics-based character animation.

\bibliographystyle{ACM-Reference-Format}
\bibliography{GenerativeControl}


\appendix

\section{Network Architecture}

\begin{table}[h]
\caption{Architecture of our Encoder and Deconv Module.}
\label{tab:architecture}
\sffamily
\small
\begin{tabular}{c}
    \hline
        \textbf{Encoder} \\
    \hline
    \begin{minipage}[t]{\linewidth}
        \begin{enumerate}[leftmargin=1.5em]
            \item Conv1D( input\_dim, 768, kernel size=(3,), stride=(1,), padding=(1,))
        
            \item 2 × Sequential(
            \begin{enumerate}
                \item Conv1D( 768, 768, kernel size=(4,), stride=(2,), padding=(1,))
                \item Resnet1D( 
                \begin{list}{}{}
                    \item 3 × ResConv1DBlock(
                    \begin{list}{}{}
                        \item (activation1): ReLU()
                        \item (conv1): Conv1D(768, 768, kernel size=(3,), stride=(1,), padding=(1,), dilation=(1,))
                        \item (activation2): ReLU()
                        \item (conv2): Conv1D(768, 768, kernel size=(1,), stride=(1,))
                    \end{list}
                    )
                \end{list}
                )
            \end{enumerate}
        \item Conv1D( 768, output\_dim, kernel size=(3,), stride=(1,), padding=(1,))
        \end{enumerate}
    \end{minipage}\\
\hspace{1em} \\
\hline
\hspace{1em}
    \textbf{Deconv} \\
\hline
\begin{minipage}[t]{\linewidth}
    \begin{enumerate}[leftmargin=1.5em]
       \item Conv1D( input\_dim, 768, kernel size=(3,), stride=(1,), padding=(1,))
   
        \item 2 × Sequential(
        \begin{enumerate}
        
        \item Resnet1D( 
        \begin{itemize}[label={}]
            \item 3 × ResConv1DBlock(
            \begin{itemize}[label={}]
                \item (activation1): ReLU()
                \item (conv1): Conv1D(768, 768, kernel size=(3,), stride=(1,), padding=(1,), dilation=(1,))
                \item (activation2): ReLU()
                \item (conv2): Conv1D(768, 768, kernel size=(1,), stride=(1,))
            \end{itemize}
            )
        \end{itemize}
        )
        \item ConvTranspose1D(768,768,kernel size=(4,),stride=(2,),padding=(1,))
    \end{enumerate}
    \item Conv1D( 768, 768, kernel size=(3,), stride=(1,), padding=(1,)) 
    \item OutputHead1D( 
        \begin{itemize}[label={}]
            \item 3 × ResConv1DBlock(
            \begin{itemize}[label={}]
                \item (activation1): ReLU()
                \item (conv1): Conv1D(768, 768, kernel size=(3,), stride=(1,), padding=(1,), dilation=(1,))
                \item (activation2): ReLU()
                \item (conv2): Conv1D(768, 768, kernel size=(1,), stride=(1,))
            \end{itemize}
            )
        \end{itemize}
        )
    \item Conv1D( 768, output\_dim, kernel size=(3,), stride=(1,), padding=(1,))
    \end{enumerate}
   \end{minipage}\\
   \hspace{1em}\\
\hline
\end{tabular}

\end{table}

\section{Respond of Claude-2}

\subsection{Example of text2motion}

\begin{itemize}
    \item \textbf{User:} \\
    Question: 'a person walks forward for a long time and kicks, then he begins to dance'
    
    \item \textbf{Claude-2:} \\
    Okay, let's see if I can interpret this sequence properly:
    
    \begin{itemize}
        \item To capture walking forward for a long time, I would use a repetitive walking sequence: \\
        297, 471, 246, 463, 463, 297, 471, 246, 463, 463, 297, 471,
        246, 463, 463, 297, 471, 246, 463, 463
        \item Then to encode the kicking motion: \\
        145, 160, 223, 117, 424, 299, 178, 166, 330, 391
        \item And finally some dancing motions like: \\
            472, 133, 132, 357, 492, 435, 217, 322, 106, 88, 279, 
            217, 387, 274, 134, 470, 175, 466, 252, 156, 134, 352, 
            436, 394, 77, 379, 297, 315, 185, 112, 428, 206, 98, 14,
            390, 390, 390, 14, 196, 395, 334

        \item The full sequence would be: \\
            297, 471, 246, 463, 463, 297, 471, 246, 463, 463, 297,
            471, 246, 463, 463, 297, 471, 246, 463, 463, 145, 160,
            223, 117, 424, 299, 178, 166, 330, 391, 472, 133, 132,
            357, 492, 435, 217, 322, 106, 88, 279, 217, 387, 274,
            134, 470, 175, 466, 252, 156, 134, 352, 436, 394, 77,
            379, 297, 315, 185, 112, 428, 206, 98, 14, 390, 390,
            390, 14, 196, 395, 334
    \end{itemize}

    Let me know if this matches the description of walking for a long time, kicking, and then beginning to dance or if I should modify the sequence in any way. Please feel free to provide any corrections to help me improve!
\end{itemize}

\subsection{Example of Abstract Task}

\begin{itemize}
    \item \textbf{User:} \\
    Question: Suppose there is a door in front of you and a key on the ground. you want to open the door and sit on the chair, how do you perform? Please describe how you get index sequences for each action.
    
    \item \textbf{Claude-2:} \\
    Here are the steps I would take to complete the task, along with the relevant motion index sequences:

    \begin{itemize}
        \item  Walk forward to reach the key on the ground: [297, 471, 246, 463, 463] I selected this "walk forward" motion sequence from the example "a person walks forward".
        \item Bend down and pick up the key: [246, 192, 408, 205, 205]
    I selected this "bend down" motion from the example "a person bends down to pick something up".
        \item Stand back up straight: [205, 198, 408] I selected this "stand up" motion from the same "bend down" example.
        \item Turn around and walk back towards the door: [463, 297, 471, 246] I reused the "walk forward" motion but in reverse order to walk back.
        \item Insert the key into the door lock and turn it: [129, 129, 433] I selected this "insert and turn" motion from the example "a person uses knife to cut object".
        \item Push the door open: [179, 179, 179]
    I selected this "push" motion from the example "a person pushes something with their right hand".
        \item Walk through the doorway: [297, 471, 246, 463] Reused the "walk forward" motion to walk through the door.
        \item Sit down on chair: [241, 241, 241] I selected this "sit down" motion from the example "a person sits down and crosses their legs".
    \end{itemize}
    
    By selecting relevant motion sequences from the examples and piecing them together in order, I was able to describe the full sequence of actions needed to open the door, enter the room, and sit on the chair inside. Let me know if you need any clarification on my approach!
\end{itemize}

\end{document}